\documentclass[journal]{IEEEtran}
\usepackage[utf8]{inputenc}
\usepackage[english]{babel}
\usepackage{makeidx}
\usepackage{graphicx}
\usepackage[hidelinks]{hyperref}
\usepackage{amsmath}
\usepackage[inline]{enumitem}
\usepackage{multirow}
\usepackage{subcaption}
\usepackage{caption}
\usepackage{color}
\usepackage{booktabs}
\usepackage{siunitx}
\usepackage{cite}
\usepackage{algpseudocode}
\usepackage{algorithmicx}
\usepackage{array}
\usepackage{float}
\usepackage{todonotes}
\usepackage[linesnumbered,ruled,vlined]{algorithm2e}
\usepackage{amsfonts}
\usepackage{threeparttable}
\usepackage{mathtools}
\usepackage{subcaption}

\algnewcommand\algorithmicforeach{\textbf{for each}}
\algdef{S}[FOR]{ForEach}[1]{\algorithmicforeach\ #1\ \algorithmicdo}

\newcommand{\T}[1]{{\color{black}{#1}}}

\newcommand{\RV}[1]{{\color{black}{#1}}}
\newcommand{\RD}[1]{{\color{blue}{#1}}}

\DeclareUnicodeCharacter{2061}{}

\begin{document}
%
\title{A Novel Incremental Learning Driven Instance Segmentation Framework to Recognize Highly Cluttered Instances of the Contraband Items}
%
%

\author{Taimur~Hassan\textsuperscript{*},~\IEEEmembership{Member,~IEEE,}
        Samet~Akcay, 
        Mohammed~Bennamoun,~\IEEEmembership{Senior Member,~IEEE,}
        Salman~Khan,
        and~Naoufel~Werghi,~\IEEEmembership{Senior~Member,~IEEE}
\thanks{\RD{© 2021 IEEE.  Personal use of this material is permitted.  Permission from IEEE must be obtained for all other uses, in any current or future media, including reprinting/republishing this material for advertising or promotional purposes, creating new collective works, for resale or redistribution to servers or lists, or reuse of any copyrighted component of this work in other works.}}
\thanks{This work is supported with a research fund from Khalifa University: Ref: CIRA-2019-047.}
\thanks{T. Hassan and N. Werghi are with the Center for Cyber-Physical Systems (C2PS), Department of Electrical Engineering and Computer Sciences, Khalifa University, Abu Dhabi, United Arab Emirates.}
\thanks{S. Akcay is with Intel R\&D UK, United Kingdom.}
\thanks{M. Bennamoun is with the Department of Computer Science and Software Engineering, The University of Western Australia, Perth, Australia.}
\thanks{S. Khan is with the Mohamed Bin Zayed University of Artificial Intelligence, Abu Dhabi, United Arab Emirates.}
\thanks{* Corresponding Author. Email: taimur.hassan@ku.ac.ae}}

\markboth{IEEE Transactions on Systems, Man, and Cybernetics: Systems, June 2020}
{Hassan \MakeLowercase{\textit{et al.}}: A Novel Incremental Learning Driven Instance Segmentation Framework to Recognize Highly Cluttered Instances of the Contraband Items}
%

\maketitle

\begin{abstract}

Screening cluttered and occluded contraband items from baggage X-ray scans is a cumbersome task even for the expert security staff. This paper presents a novel strategy that extends a conventional encoder-decoder architecture to perform instance-aware segmentation and extract merged instances of contraband items without using any additional sub-network or an object detector. The encoder-decoder network first performs conventional semantic segmentation and retrieves cluttered baggage items. The model then incrementally evolves during training to recognize individual instances using significantly reduced training batches. To avoid catastrophic forgetting, a novel objective function minimizes the network loss in each iteration by retaining the previously acquired knowledge while learning new class representations and resolving their complex structural inter-dependencies through Bayesian inference. A thorough evaluation of our framework on two publicly available X-ray datasets shows that it outperforms state-of-the-art methods, especially within the challenging cluttered scenarios, while achieving an optimal trade-off between detection accuracy and efficiency.

\end{abstract}

\begin{IEEEkeywords}
Baggage X-ray Scans, Semantic Segmentation, Instance Segmentation, Incremental Learning.
\end{IEEEkeywords}

\IEEEpeerreviewmaketitle

\section{Introduction}
\label{sec:intro}
\IEEEPARstart{T}{he} inspection of passenger's baggage, packages, and containers with X-ray scanners is nowadays a part of the standard checking measures in airports and any other public place where safety and security are of significant concern. 
This screening process is cumbersome, requiring the relentless attention of a human expert. Furthermore, it's vulnerable to human errors caused due to exhausting work schedules, lack of experience, and the concealed nature of the contraband items.  
Although object detection in color images has been a rigorously researched topic, its applicability to X-ray-based threat detection is somewhat limited. The primary reason is the remarkably different X-ray imagery characteristics, where texture and appearance details are scarce compared to regular color images. An adequate system for such a critical application is expected to detect objects under high occlusions, in cluttered scenes, with large view-point variations and limited amounts of contraband data.
\begin{figure}[t]
\centering
\includegraphics[width=1\linewidth]{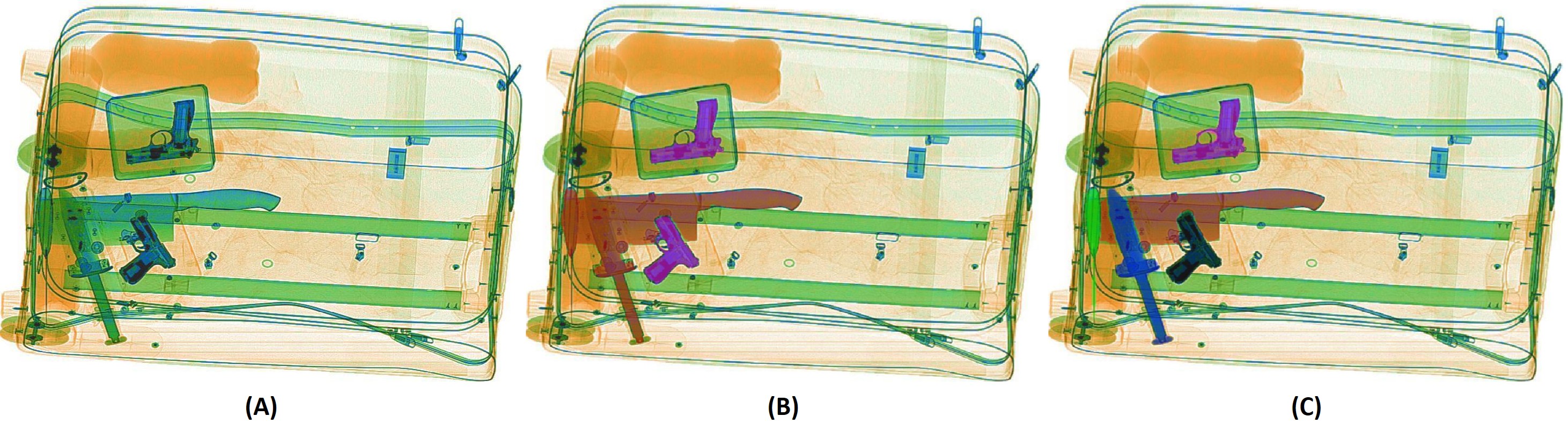}
\caption{\small \T{(A) An original X-ray scan from the SIXray dataset \cite{miao2019sixray}, (B) conventional semantic segmentation, and  (C) instance-aware segmentation.} }
\label{fig:semanticEg}
\end{figure}
Many researchers have developed supervised and unsupervised screening systems for detecting contraband items in X-ray images in response to these challenges. The most recent wave of these efforts employed deep learning models, particularly one-staged and two-staged object detectors such as RetinaNet \cite{retinanet}, YOLO \cite{yolo}, and Faster R-CNN \cite{fasterrcnn}. While these systems showed remarkable capacity for detecting isolated objects, their performance degrades in recognizing extremely cluttered, occluded, and overlapping items \cite{akcay2018using, gaus2019evaluation}. 
\RV{Semantic segmentation models, due to their pixel-level recognition ability, can extract the extremely occluded contraband items from X-ray baggage scans \cite{Hassan2020ACCV}. With the integration of object context in the pixel classification, they have more potential to improve the threat detection accuracy \cite{an2019}. By leveraging this capacity, some of the initial attempts employed the encoder-decoder-encoder topology for detecting suspicious items as anomalies \cite{akcay2018ganomaly}. However, semantic segmentation networks have an inherent limitation of detecting the individual instances of the overlapping items. For example, in Figure \ref{fig:semanticEg} (B), we can see that how a semantic segmentation network cannot recognize the overlapping \textit{kitchen knife} and \textit{chopper} individually. In such scenarios, these networks output only a single blob in which the information about individual item instances is lost. 
Detecting individual instances of the same threat category is, in fact, desirable in cases where we need to identify and locate each instance precisely \T{(see the example in Figure \ref{fig:semanticEg}-C, where the \textit{kitchen knife} and the \textit{chopper} instances have been extracted separately).} Also, identifying individual items’ instances is vital in aviation baggage screening as some instances of the items are legal to carry within the baggage, whereas some instances are prohibited. For example, passengers can carry certain \textit{drugs} and \textit{bottles} in their luggage, but \textit{addictive drugs} and \textit{alcoholic drinks} are banned at airports \cite{EU}. 
}
Towards this end, Gaus et al. \cite{gaus2019evaluation, gaus2019evaluating} introduced an instance segmentation approach in their baggage threat detection system using Mask R-CNN \cite{maskrcnn}. However, the authors realized that conventional instance segmentation network requires extensive ground truth labeling and exhaustive training efforts, especially for the large-scale datasets, and there is a need to develop a framework that can effectively perform instance-aware segmentation to recognize the cluttered contraband items from the baggage X-ray imagery via incremental few-shot training. 

\section{Related Work} \label{sec:relatedWorks}
\noindent Existing solutions for contraband item detection based on X-ray imagery can be classified as traditional machine learning and deep learning methods. In this section, we shed light on the main approaches, and we refer the reader to the work of \cite{Mery2017TMSC}, and \cite{ackay2020} for a detailed survey. \RV{In addition to this, this section also explores the recent advances in incremental learning to perform classification and segmentation tasks.}

\noindent \textbf{A. Conventional Machine Learning Methods:}
\noindent The initial methods developed for screening contraband items employ conventional machine learning. These solutions are either based on classification \cite{bastan2013BMVC}, detection \cite{bastan2015} or the segmentation approaches \cite{heitz2010}. Bastan et al. \cite{bastan2011} used SURF features with Bag of Words (BoW) to identify suspicious objects. Instead of SURF, Kundegorski et al. \cite{kundegorski2016} utilized FAST-SURF with BoW to classify prohibited baggage items. Other works involve Adaptive Sparse Representation \cite{mery2016}, and Adapted Implicit Shape Model \cite{riffo2015automated} to detect contraband data. 
Apart from this, Mery et al. \cite{mery2016} developed a framework that computes 3D feature points through the structure from motion and uses these features to classify contraband items from the X-ray imagery. 

\noindent \textbf{B. Deep Learning Methods:}
\RV{\noindent The most recent deep learning methods can be categorized either as supervised detection and segmentation approaches or as unsupervised adversarial learning schemes.}

\noindent \textbf{\textit{1. Supervised Detection Strategies:}}
The majority of deep contraband item detection frameworks 
utilizes one-staged or two-staged object detectors such as YOLOv2 \cite{yolov2}, RetinaNet \cite{retinanet} and Faster R-CNN \cite{fasterrcnn}. Moreover, researchers have also utilized pre-trained models for the object classification within baggage X-ray scans \cite{jaccard2017, gaus2019evaluation}. 
Zou et al. \cite{_41} utilized YOLOv2 \cite{yolov2} to detect \textit{scissors}, \textit{knives} and \textit{bottles} from their local 1,104 synthetic X-ray images.  
 Miao et al. \cite{miao2019sixray} released the largest security inspection X-ray dataset (SIXray) that contains highly occluded and overlapping instances of contraband items such as \textit{guns}, \textit{knives}, \textit{wrenches}, \textit{pliers}, \textit{scissors} and \textit{hammers}. 
Furthermore, they presented a framework dubbed class-balanced hierarchical refinement (CHR) to recognize contraband items from the SIXray \cite{miao2019sixray} dataset. 
More recently, Hassan et al. \cite{hassan2019} presented Cascaded Structure Tensor (CST) framework that generates contours-driven bounding boxes of potentially prohibited items which are then classified using ResNet\textsubscript{50} \cite{he2016deep}. 

\RV{\noindent \textbf{\textit{2. Supervised Segmentation Approaches:}}
Apart from solving the baggage threat recognition problem via deep object detection methods, many researchers utilized semantic and instance segmentation as a tool to effectively recognize suspicious baggage content \cite{an2019, gaus2019evaluation}. It is essential to note here that although we can fine-tune standard encoder-decoder networks for a large variety of semantic segmentation tasks, specific applications would be best be approached with customized models \cite{liang_eccv_2018}.
For example, to cope with object size variation and camera view changes in traffic and surveillance applications,  Akilan et al. \cite{akilan2020} proposed integrating residual feature fusions at early, middle and late stages in the encoder-decoder architecture (dubbed MvRF-CNN \cite{akilan2020}). Similarly, driven by achieving the optimal trade-off between the segmentation accuracy and the computational model complexity, Wang et al. \cite{wang_cvpr2020} coupled an encoder-decoder model and super-resolution construction scheme. Similarly, a multi-task attention network is proposed in \cite{wang2020MAN} that coupled handcrafted features pipeline and an attention network to segment the object of interest \cite{wang2020MAN}. Also, an adversarial domain adaptation scheme is proposed in \cite{wang2019ADA} that employs a detection and segmentation (DS) model along with domain classifiers to learn target domain labels from the source domain synthetic data in a weakly supervised manner. In addition to this, Hassan et al. \cite{Hassan2020ACCV} proposed a contour instance segmentation strategy that segments the suspicious baggage content by analyzing the strength of the variation within their contours \cite{Hassan2020ACCV}.
}

\noindent \textbf{\textit{3. Unsupervised Adversarial Learning:}}
Apart from supervised learning frameworks for detecting contraband items, Akcay et al. proposed GANomaly \cite{akcay2018ganomaly}, and Skip-GANomaly \cite{akccay2019skip} to derive the latent space representation of the contraband items in an adversarial manner to recognize them as anomalies within the baggage X-ray scans. 


\noindent \textbf{C. Incremental Learning Strategies:}
\noindent Incremental learning schemes have gained immense popularity in the context of deep learning for overcoming the need for excessive computational burden in re-training models with large-scale data, which might be difficult to obtain and prepare.
However, developing an incremental learning scheme that overcomes catastrophic forgetting (the tendency of a deep learning model to drastically forget the prior knowledge while learning about new information) is also challenging.  
To address this, many researchers have proposed schemes involving knowledge distillation \cite{kd}, 
gating \cite{gate}, 
and indefinitely long term learning (iCaRL) \cite{icarl}. Furthermore, Tian et al. \cite{crd} exploited the fact that knowledge representations exhibit complex relationships that cannot be learned through objective functions that assume independence of events. 
Cho et al. \cite{cho} advocated that good performing teachers do not necessarily produce good students due to the student network's limited capacity to cope with the teacher's growing knowledge. 
Lopez-Paz et al. \cite{lopez} proposed the Gradient Episodic Memory (GEM) scheme, which uses episodic memories to hold a small set of examples from the prior learned tasks to avoid catastrophic forgetting. 
Apart from this, researchers have also proposed distillation-driven incremental learning strategies for performing the semantic segmentation tasks \cite{kdils}.

\begin{figure*}[t]  
\centering
\includegraphics[width=1\linewidth]{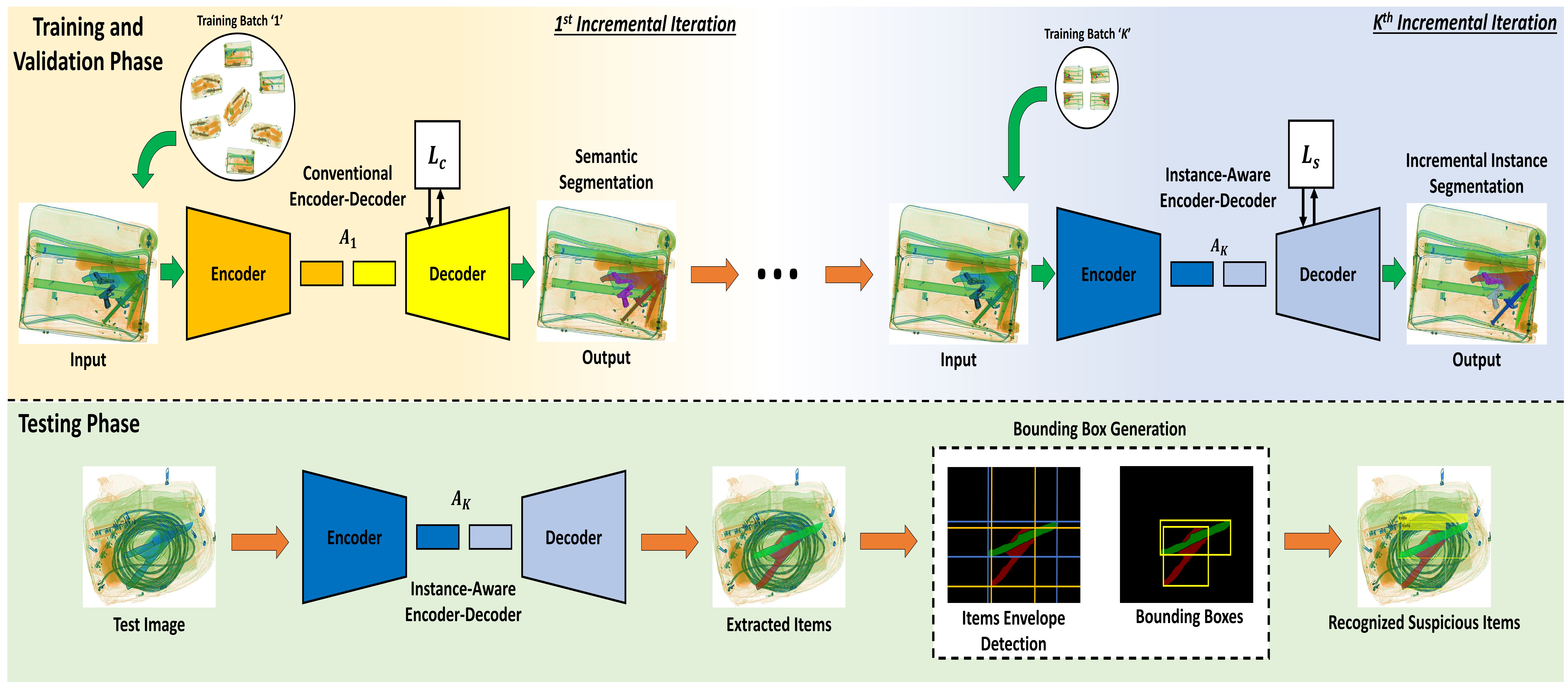}
\caption{ \small \T{Block diagram of the proposed framework. We trained the proposed model incrementally to recognize cluttered instances of the contraband items. At each iteration, $k=1,...,K$, the number of item instances that the system can recognize is incremented by one.
At the inference stage, the model $A_{K}$ (incrementally trained till $K^{th}$ iteration to recognize up to $K$ overlapped instances) is used for the instance-aware segmentation of the cluttered contraband items. More on the training details are in  Section (\ref{sec:training}-B).}}
\centering
\label{fig:seg}
\end{figure*}

\RV{
\noindent \textbf{D. Limitations of Existing Work:}
\noindent The main limitations of the existing approaches are their inadequate validation on single datasets or their application to simplistic scenarios within a very constrained environment. For instance, the problem of robustly detecting cluttered, occluded, and overlapping contraband items from the highly imbalanced datasets is still an open question to be addressed. The approaches proposed in \cite{miao2019sixray}, \cite{hassan2019} and \cite{Hassan2020ACCV} handles such cases. However, they produce either low detection performance \cite{miao2019sixray} or are subject to parameter tuning \cite{hassan2019}. 
Apart from this, researchers have also utilized semantic segmentation networks to recognize suspicious baggage content via X-ray imagery \cite{an2019}. Such models have improved the performance of the threat detection frameworks. However, they cannot distinguish between cluttered and overlapping instances of the same items (e.g., a \textit{knife} overlaid on another \textit{knife} as shown in Figure \ref{fig:semanticEg}-B), which is often desirable in aviation screening, and for such cases, the semantic segmentation networks output a single blob of pixels representing only a single class label. To cater this, Gaus et al. \cite{gaus2019evaluation} introduced the usage of Mask R-CNN \cite{maskrcnn} for baggage threat detection. However, the Mask R-CNN-based threat detection system presents limitations in extracting the cluttered contraband items because it relies on the region-based proposals that fail to detect cluttered objects correctly \cite{gaus2019evaluation}. This limitation of Mask R-CNN \cite{maskrcnn} and other instance-aware segmentation networks will be further evidenced when employed in complex datasets such as SIXray \cite{miao2019sixray}, as described in Section \ref{sec:results}. Moreover, other approaches utilized encoder-decoder architectures and fully convolutional networks coupled with classification sub-networks or region of interest (ROI) voting to recognize multiple objects instances individually \cite{an2019,multipathnet}. However, these frameworks also produce a poor trade-off between detection accuracy and efficiency. On the other hand, instance segmentation frameworks require extensive bounding box and mask-level annotations \cite{Hassan2020ACCV}, which are reasonably hectic, and resource-demanding to procure, especially for large-scale datasets, such as SIXray \cite{miao2019sixray}. Also, training such networks requires an excessive amount of memory and computational resources. To alleviate these problems, we propose an incremental learning-driven instance-aware segmentation approach, as discussed below. 
}

\noindent \textbf{E. Contributions:}
\noindent This paper proposes a novel scheme that utilizes incremental learning to make conventional semantic segmentation models instance-aware. The proposed method is simple and exhibits modest training efforts by requiring only a small batch of training samples to add more instances of a given suspicious item class. This strategy bypasses hectic annotation workflows as are necessary for training traditional instance segmentation frameworks while overcoming the excessive memory and computational requirements. The proposed framework also avoids catastrophic forgetting through an instance segmentation objective function that minimizes the network loss to retain knowledge about the previously learned classes while understanding new class representations and resolving their complex inter-dependencies. The unique characteristics of the proposed system are:

\begin{itemize}[leftmargin=*]
    \item A novel approach that extends conventional encoder-decoder networks to recognize individual instances of the contraband items from the X-ray scans.

    \item No requirement for an additional object detector, classification sub-network, or ROI voting to perform instance-aware segmentation.

    \item An incremental learning-driven instance segmentation framework that discriminates the overlapping and isolated suspicious item instances with only a few training examples.
    
    \item Robust to catastrophic forgetting due to its ability to resolve complex inter-dependencies between already learned and newly added suspicious items categories.
    
\end{itemize}

\noindent The rest of the paper is organized as follows: 
Section \ref{sec:proposedFramework} discusses the proposed system. Section \ref{sec:expSetup} enlists the experimental plan. Section \ref{sec:results} presents the experimental results. Section \ref{sec:discussion} contains a detailed discussion on the performance of the proposed system and Section \ref{sec:conclusion} presents concluding remarks.

\section{Proposed Framework} \label{sec:proposedFramework}
\noindent Figure \ref{fig:seg} depicts the block diagram of the proposed framework. This framework trains an encoder-decoder model to recognize up to $K$ isolated and overlapped instances of a given class incrementally in  $K$ iterations. The first iteration reflects the ordinary semantic segmentation to extract different contraband items from the baggage X-ray scans. For this, we train the first instance of encoder-decoder dubbed $A_{1}$ on a relatively large set of training images. 
Afterward, we make the encoder-decoder model instance-aware in each iteration by exposing it to the small training batches. For example, in the $k^{th}$ iteration, we make the encoder-decoder $A_{k}$ to recognize up to $k$ instances of the same item by providing a different set of corresponding images. The final instance-aware segmentation model is obtained at the iteration $K$. In this process, the model is immunized to catastrophic forgetting by analyzing the complex relationships between previously learned and newly added suspicious item categories through the proposed loss function (see Eq. \ref{equ:loss}). Before exposing the details of our approach, we provide a brief overview of the incremental learning paradigm in the next section for completeness. 

\noindent \textbf{A. Incremental Learning:}
\noindent In a conventional incremental learning paradigm, the model is trained iteratively. At each iteration $k$, it performs $C_W$-class
segmentation (or classification) task where $C_W$ denotes the number of classes in the current iteration  $k$. To learn this task, the model is given a set of  $\mathcal{D}$ training samples such that $\mathcal{D}=\{\mathcal{D}_o, \mathcal{D}_n\}$,
where $\mathcal{D}_o$ denotes the samples of old classes $W_o$, learned from iteration \textit{1} to  ($k-1$), and $\mathcal{D}_n$ denotes the samples of newly added classes ($W_n$) to be learned in the current iteration $k$.  The cumulative list of all the classes (both $W_o$ and $W_n$) is represented by $W$, i.e., $W = \{W_o, W_n \}$. The network is also fed with the ground truth $t=\{t^o, t^n\}$ of these training samples where $t^o$ and  $t^n$  denote the ground truth for the samples of old classes and the new classes, respectively. $t$ is normally represented in a one-hot encoding vector notation \cite{oneHotEncoding}. These training samples are passed as an input to the network for which it generates the output logits $l$ in the last layer such that $l=v f + \gamma$, where $f$ represents the feature vector, $v$ represents the layer weights, and $\gamma$ denotes the biasing factor. The logits $l=\{l^o, l^n\}$  are the concatenation of the old logits $l^o$ and the new logits $l^n$, generated from training the old classes and the newly added classes, respectively.
These logits are then passed through the activation function (usually softmax) in the final layer of the CNN model to generate the final class probabilities, i.e., $ p(l_{i,j}) = \frac{\exp{(l_{i,j})}}{\sum_{r=0}^{C_W-1} \exp{(l_{i,r})}}$, where $p(l_{i,j})$ denotes the probability of the  $i^{th}$ training sample being part of the $j^{th}$ class. 
$p(l_{i,j})$ in the above definition is known as a hard class probability of the logit $l_{i,j}$.  Hard probabilities are generally recommended in traditional classification or segmentation task because they clearly discriminate the most expected class out of the rest. But in incremental learning, logits are scaled  using the temperature constant ($\tau$) to generate the soft target probabilities, i.e., $ p(l_{i,j}^\tau) = \frac{\exp(l_{i,j}^\tau)}{\sum_{r=0}^{C_W-1} exp(l_{i,r}^\tau)} $, where $l_{i,j}^\tau = l_{i,j}/\tau$. Here, $\tau$ is used to increase the degree of relaxation of the soft label by reducing the disparities between classes probabilities. Practically, it is a hyper-parameter which is tuned for the sake of obtaining a better performing model \cite{ILSurvey}.

\begin{figure*}[htb]
    \centering
    \includegraphics[width=0.8\linewidth]{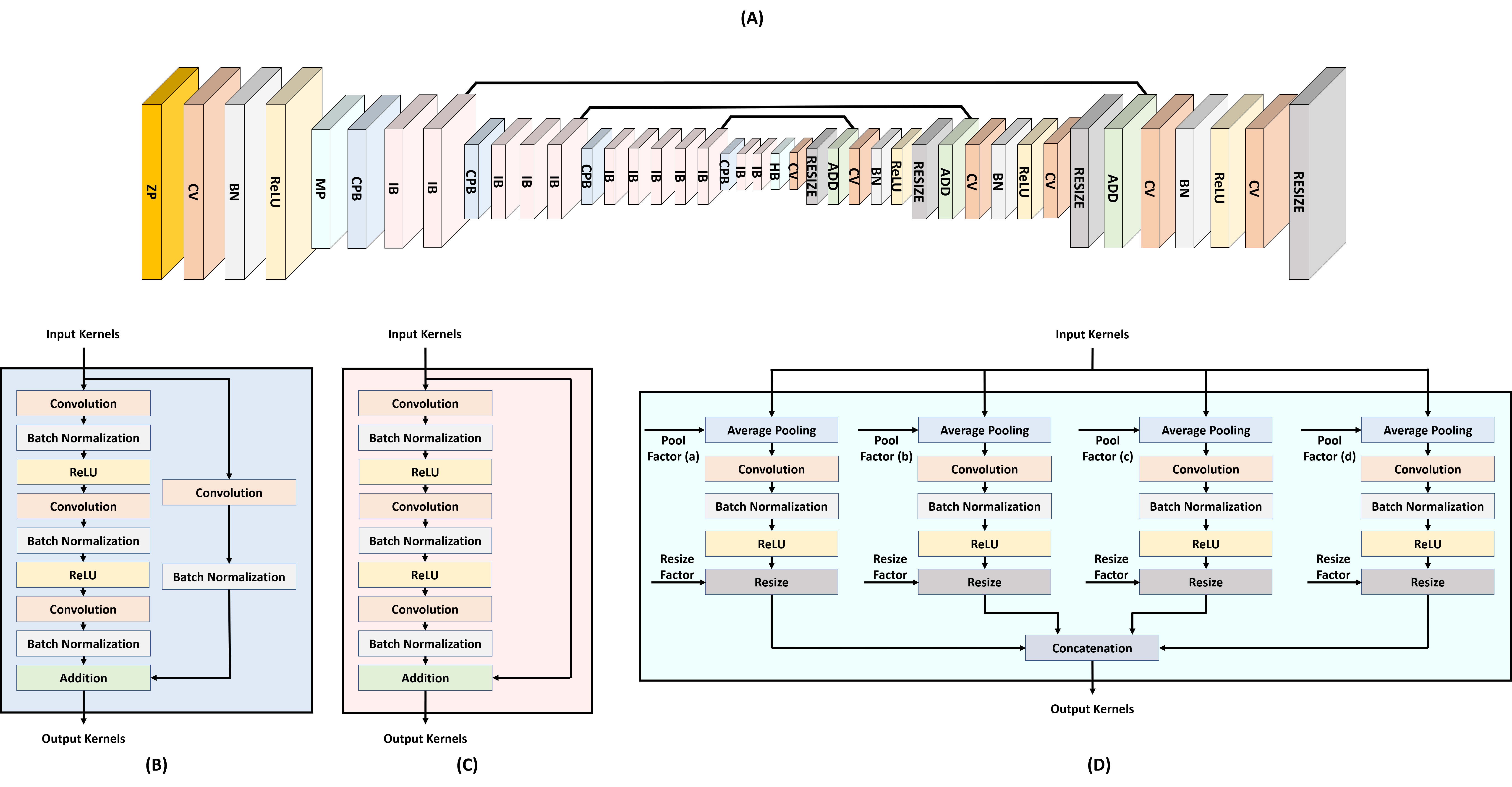}
    \caption{ {\small \color{black}{(A) CIE-Net architecture, (B) contextual preservation block (CPB), (C) identity block (IB), (D) hierarchical block (HB). Moreover, CV, BN, MP, and ZP in (A) denote the convolution, batch normalization, max pooling, and the zero-padding layer, respectively.}
    }} 
    \label{fig:cienet}
\end{figure*}

\noindent \textbf{B. Semantic Segmentation:}
\T{\noindent The first iteration of the proposed framework relates to semantic segmentation, where we train the proposed contraband items extraction network (CIE-Net) 
to extract different contraband items from the baggage X-ray images.
}
\T{\noindent The prime objective of designing the proposed CIE-Net is to accurately extract the contraband items and their instances, even in overly cluttered scenarios. We utilize convolutional blocks (with ReLU activations and batch normalizations) to preserve coarser feature representations of the contraband items while simultaneously retaining their geometrical shapes through finer edge information. The blocks follow a hierarchical design to yield multi-scale representations of threat objects for superior mask-level extraction. Furthermore, we implant novel identity blocks within the encoder topology of the CIE-Net that further aids in preserving the object's geometrical characteristics regardless of the amount of clutter. The optimal values for the number of filters and kernel sizes are determined empirically after analyzing the similarly designed frameworks like PSPNet \cite{zhao2017pyramid}, and ResNet \cite{he2016deep} to craft out the optimal design schematics for the CIE-Net.}

\T{\noindent The detailed architecture of CIE-Net is illustrated in Figure \ref{fig:cienet}. Here, we can observe that the CIE-Net consists of an asymmetric encoder-decoder topology. The desired objects' contextual and geometrical features are preserved through the contextual preservation blocks (CPB), composed of cascaded convolution and batch normalization operations. CPB ensures that the network learns to discriminate the similar textured contraband items (even the cluttered ones) by tuning the network weights based upon categorical cross-entropy loss function ($L_c$) in the first iteration, and the proposed instance segmentation loss function ($L_s$) in the rest of the iterations. Moreover, to ensure that the network retains the finer shape representations of the contraband items, dedicated identity blocks (inspired by ResNet \cite{he2016deep} scheme) have been added in the encoder part, where the finer representations (of the suspicious items) are fused with the decoder end via residual triggered skip-connections. Inspired by PSPNet \cite{zhao2017pyramid}, we also employ a custom hierarchical block (HB) to improve the performance of the CIE-Net further. HB uses variable pooling factors (determined empirically) to generate the multi-scale feature representations from the latent vector space to recognize the cluttered contraband items and their instances. The hierarchical decomposition and pooling factors are determined empirically to obtain the optimal contraband item extraction performance on grayscale and colored baggage X-ray scans.

\noindent Like the proposed framework, the MvRF-CNN \cite{akilan2020} also preserves the desired objects' geometrical information by fusing feature representations obtained across various network depths in a residual manner \cite{akilan2020}. Similarly, to achieve better geometrical characteristics of the desired objects, the framework proposed in \cite{wang_cvpr2020} couples a segmentation encoder-decoder model with the super-resolution construction scheme where the fine-grained structural features are derived through the affinity maps \cite{wang_cvpr2020}. To have a precise idea of how the above model works to detect baggage threats from security X-ray scans, we evaluated them both on the GDXray \cite{mery2015gdxray}, SIXray \cite{miao2019sixray}, and the combined datasets. We also compared these scheme's performance with the proposed incremental instance segmentation framework (please see Table \ref{tab:iou} for more details).
}

\noindent In the first incremental training iteration, CIE-Net optimizes the ${L}_{c}$ function to discriminate between normal and suspicious items (in a semantic segmentation fashion): 
\begin{equation}
{L}_{c} = -\frac{1}{N_t}\sum\limits_{i=0}^{N_t-1}\sum\limits_{j=0}^{C_W-1} t_{i,j}\log(p(l_{i,j})),
\end{equation}
where $C_W$ denotes the total number of classes for the current iteration, $N_t$ represents the total number of samples in the training batch, for the current iteration, $t_{i,j}$ is a binary value  telling whether the  $i^{th}$ sample represents the $j^{th}$ class or not, and $p(l_{i,j})$ is the probability of the logit ($l_{i,j}$) of $i^{th}$ sample for the $j^{th}$ class. 

\noindent Here, we also want to highlight that the semantic segmentation network extracts isolated and merged suspicious items from the baggage X-ray scans in the first iteration. However, the network cannot differentiate between multiple instances of the same item (e.g., two or more \textit{knives} or \textit{guns} in a single scan, whether they are isolated or merged).

\noindent \textbf{C. Incremental Instance Segmentation:}
\noindent We propose a novel instance segmentation framework that utilizes incremental learning to make conventional semantic segmentation networks instance-aware. 
Most of the instance-aware segmentation models employ object detectors, ROI voting, or separate classification sub-networks. However, such implications require additional overheads for preparing large-scale training data and excessive memory requirements. Contrary to this, our proposed scheme makes conventional encoder-decoder models instance-aware without needing any additional resources. Thanks to the incremental adaptation strategy, only a small-scale training batch is required in each iteration to learn about multiple item instances in each scan, which drastically reduces the memory and computational requirements compared to the fine-tuning approaches. Furthermore, our framework has an in-built capacity to resist catastrophic forgetting through the proposed incorporation of the mutual information loss function, which analyzes the complex inter-dependencies between prior knowledge and newly learned information through Bayesian inference. 

\noindent \textbf{\textit{1. Instance Segmentation Loss Function:}} 
For instance-aware segmentation, we propose the following loss function.
\begin{equation}
L_{s} = \alpha_1 L_{n} + \alpha_2 L_{o} + \alpha_3 L_{mi}, 
\label{equ:loss}
\end{equation}
where $\alpha_{\{1, 2, 3\}}$ denote the loss weights (determined empirically to be 0.2, 0.3, and 0.5).
$L_{n}$ minimizes the network loss for learning new instance categories, and $L_{o}$ minimizes the distillation loss for retaining the prior learned knowledge (about segmenting the suspicious baggage items). Both $L_{n}$ and $L_{o}$ are widely used in continual learning frameworks to avoid catastrophic forgetting \cite{ILSurvey}. In the proposed framework, $L_{o}$ is calculated through categorical cross-entropy loss, while $L_n$ is calculated through KL divergence loss, as shown below:

\vspace{-.2cm}

\begin{equation}
L_{o} = -\frac{1}{N_{t_o}}\sum\limits_{i=0}^{N_{t_o}-1}\sum\limits_{j=0}^{C_{W_o}-1} t_{i,j}^{o} \log(p(l_{i,j}^{o,\tau})),
\label{eq:Lo}
\end{equation}

\vspace{-.5cm}

\begin{equation}
\begin{split}
L_{n} = \frac{1}{N_{t_n}}\sum\limits_{i=0}^{N_{t_n}-1}\sum\limits_{j=0}^{C_{W_n}-1} q(t_{i,j}^{n}) \log \left(\frac{q(t_{i,j}^{n})}{p(l_{i,j}^{n,\tau})}\right),
\end{split}
\label{eq:Ln}
\end{equation}
where $N_{t_o}$ and  $C_{W_o}$  denote, respectively, the number of old training samples and the number of old classes (added in 1 to $k-1$ iterations). $N_{t_n}$ and $C_{W_n}$ denote, respectively, 
the number of new training samples and the number of newly added categories (in the current $k^{th}$ iteration). $t^o_{i,j}$ and $t^n_{i,j}$ represent, respectively, the ground truth for the training samples of the old and the new classes. $p(l_{i,j}^{o,\tau})$  is the predicted distribution of the scaled logits generated through the training samples of the old classes. $q(t_{i,j}^{n})$ is the actual distribution generated from the true labels of the newly added classes, $p(l_{i,j}^{n,\tau})$  represents the predicted distribution of the scaled logits generated through the training samples of the new classes. 

\noindent $L_{mi}$ in Eq. (\ref{equ:loss}) is the new proposed loss term, which we introduce to account for the inter-dependencies between old knowledge and newly learned information in our problem. More about the rationale and the description of this loss term is given in the next sub-section. 

\noindent \textbf{\textit{2. Mutual Information Loss Function}}
\T{
The mutual information loss function ($L_{mi}$) is based on the Bayesian inference that exploits the complex inter-dependencies between previously learned class representations (in iteration 1 to $k-1$) through their respective training examples and the examples related to the newly stacked classes (in the current iteration $k$).  $L_{mi}$ is expressed as follows:

\vspace{-.4cm}

\begin{equation}
\begin{split}
L_{mi}=-\frac{1}{N_{t_o}}\sum\limits_{i=0}^{N_{t_o}-1}\sum\limits_{j=0}^{C_{W_o}-1} t^{o}_{i,j} \log(p(w_j  | l_{i,j}^{o,\tau},l_{i,j}^{n,\tau})) ,
\end{split}
\label{eq:label8}
\end{equation}
where $N_{t_o}$ denotes the total number of training examples, $C_{W_o}$ denotes the total number of old classes ($W_o$), and $t^{o}_{i,j}$ the  ground  truth  for  the training samples representing the previously added classes (in iterations 1 to $k-1$). The posterior probability  $p(w_j|l_{i,j}^{o,\tau},l_{i,j}^{n,\tau})$ is defined as:
\begin{equation}
 p(w_j|l^{o,\tau}_{i,j},l^{n,\tau}_{i,j}) =  \frac{p(l^{o,\tau}_{i,j},l^{n,\tau}_{i,j} | w_j)\times p(w_j)}{\sum_{k=0}^{C_{W_o}-1}p(l^{o,\tau}_{i,k},l^{n,\tau}_{i,k}  |w_k) p(w_k)} .
    \label{eq:label18}
\end{equation}
\T{\noindent It should be noted here that the evidence $\sum_{k=0}^{C_{W_o}-1}p(l^{o,\tau}_{i,k},l^{n,\tau}_{i,k}  |w_k) p(w_k)$ in Eq. (\ref{eq:label18}) is an optional term because it only normalizes the probability distribution $p(w_j|l^{o,\tau}_{i,j},l^{n,\tau}_{i,j})$, so that the sum of probabilities for all the outcomes is 1. 
}

\noindent The rationale of encompassing $L_{mi}$ stems from the fact that older class representations (learned across the $k-1$ iterations) and the newly learned categories (in the  $k^{th}$ iteration) are non-mutually exclusive. For example, a network trained to extract \textit{knives} (particularly \textit{kitchen knives}) in the first iteration should be aware of the contextual similarity between \textit{kitchen knives} and \textit{choppers} (which it learns in the second iteration) since both of them are different type of \textit{knives}. 

\noindent To the best of our knowledge, all the knowledge distillation and incremental learning solutions handle catastrophic forgetting by separately minimizing the network loss involved in learning the new tasks and maintaining the prior learned knowledge inferred from the previous model (or teacher) instance. But the frameworks, trained using these loss functions assume that both older and newly added class representations are independent of each other, leading towards compromised performance, especially in those scenarios when the incrementally learned information highly correlates with one another. In our approach, the additional loss function ($L_{mi}$) integrates the relationship between prior learned and recently stacked classes through their training examples and exploits it via Bayesian inference to maximize the capacity of the incremental learning process of differentiating contraband item instances.}

\noindent \textbf{D. Bounding Box Generation:} \label{bboxGeneration}
\noindent After extracting the suspicious items from the candidate scan, the bounding box for each extracted item \T{($\zeta$) is generated through a simple yet very effective scheme. We iterate over the mask of each extracted contraband item ($\zeta$) within the candidate scan, where for each mask, we find its minimum and maximum row value. The minimum row value represents the minimum row index within the candidate scan, where the mask value is one. Similarly, the maximum row value represents the maximum row index (within the candidate scan), where the mask is 1. Afterward, we take the image transpose and repeat the same process to get the minimum and maximum column index required to generate (and fit) the bounding box. The mathematical expression of the whole scheme is as follows:

\vspace{-.5cm}

\begin{equation}
\begin{aligned}
    <y_{min}, y_{max}> = <\underset{0 \leq u \leq M-1}{argmin}(\zeta_u), \underset{0 \leq v \leq N-1}{argmax}(\zeta_v)> ,
\end{aligned}
\end{equation}

\vspace{-.5cm}

\begin{equation}
\begin{aligned}
    <x_{min}, x_{max}> = <\underset{0 \leq u \leq M-1}{argmin}(\zeta_u^T), \underset{0 \leq v \leq N-1}{argmax}(\zeta_v^T)> ,
\end{aligned}
\end{equation}

\vspace{-0.5cm}

\begin{equation}
\beta_b = [x_{min}, y_{min}, x_{max}-x_{min}, y_{max}-y_{min}],
\end{equation}

\noindent where $u, v \in \mathbb{W}$, $M$ and $N$ denotes the width and height of $\zeta$, respectively, and $\beta_b$ denotes the bounding box of the candidate contraband item (generated via its extracted mask).
}
\section{Experimental Setup} \label{sec:expSetup}
\noindent This section reports the datasets, the training details, and the evaluation metrics (used in the evaluation and also in the comparative study). 

\noindent \textbf{A. Datasets:} 
\noindent We evaluated the proposed framework on publicly available GDXray \cite{mery2015gdxray}, SIXray \cite{miao2019sixray}, and the combined dataset (containing the scans from both GDXray \cite{mery2015gdxray} and SIXray \cite{miao2019sixray} datasets). We report the detailed description of these datasets in the supplementary material (and in the source code repository\footnote{\label{note1} \noindent \color{black}{The complete source code and its documentation is available at: \url{https://github.com/taimurhassan/inc-inst-seg}.}}) due to space constraints.

\noindent \textbf{\T{B. Incremental Training Details:}} \label{sec:training}
\T{
\noindent To incrementally train the proposed framework on the GDXray \cite{mery2015gdxray} dataset, we used a total of 788 scans (400 scans for extracting originally identified suspicious items and 388 scans for the locally identified items).
\noindent However, for the SIXray \cite{miao2019sixray} dataset, we used 80\% of the scans for training and 20\% for evaluation as per the dataset standard \cite{miao2019sixray}. Note that the number of incremental training iterations depends on the number of cluttered item instances within each dataset. In the combined dataset, we have a total of 1,067,381 scans in which 27,750 scans (13,663 positives and 14,087 negatives) were used for training purposes, and the rest of 1,039,631 scans were used in the evaluations. Such a training split also ensures assessing the resistance of the proposed framework against class imbalance. 

\noindent Moreover, in the first training iteration, we constrain the network with the $L_c$ loss function to recognize different contraband items. Here, the proposed model performs conventional semantic segmentation to extract, for example, a \textit{gun} and a \textit{knife} contained within the candidate scan. However, it should be noted that the semantic segmentation model cannot recognize the \RV{overlapping} instances of the same item, i.e., a \textit{gun} overlaid on another \textit{gun}. In such scenarios, the semantic segmentation models will output a single blob of \textit{gun}-labeled pixels. 

\noindent To accurately recognize the individual overlapped instances of contraband items (e.g., two \RV{overlapping} \textit{guns}), we further train our model iteratively. In each incremental iteration, we stack new classes, representing individual instances of the contraband items. Through their respective training examples, we re-tune the proposed model to make it instance-aware. For example, in the second iteration, we train the proposed model to recognize at most two overlapped instances of any suspicious item \RV{(e.g., two instances of \textit{guns}, two instances of \textit{knives} etc.)} by stacking two additional classes representing \textit{gun} and \textit{knife} instance. 
We, therefore, feed the network with a small batch of training examples (containing at most two overlapping instances), where the two overlapping suspicious items (e.g., two overlapping \textit{guns}) are marked with two different class labels in the ground truth. The same process is repeated across all the iterations until we obtain \textit{K}-instance aware segmentation model where $K$ denotes the maximum overlapping instances of the same item within the dataset. In addition to passing training examples representing the newly stacked classes, we also pass a few examples representing the previous classes (added in the iterations 1 to $k-1$).
The set of samples used to train the proposed model at each iteration is significantly lesser than the amount of data that is required by its competitors \cite{Hassan2020ACCV, hassan2019, akcay2018using, miao2019sixray}, i.e., it only uses around 20\% of the total training data (defined as per the dataset standard), wherein each increment,  about 10\% examples are added to retain the knowledge of the previously learned categories.

\noindent The training is conducted on a machine with an Intel Core i7-9750H@2.6 GHz processor and 32 GB RAM with a single NVIDIA RTX 2080 Max-Q GPU, cuDNN v7.5, and a CUDA Toolkit v11.0.221. The CIE-Net is implemented using TensorFlow 2.1.0 with Keras 2.3.0 on the Anaconda platform using Python 3.7.9. In the first iteration, the training consisted of 20 epochs, whereas the subsequent iterations took ten epochs with ADADELTA \cite{Zeiler2012ADADELTA} optimizer. Moreover, the exact number of learnable and non-learnable parameters in CIE-Net varies in each iteration. Still, on average, they are roughly around 31.4M and 61.3K, respectively. The detailed model architecture is available in the codebase repository\textsuperscript{\ref{note1}}.

\noindent We also tested the proposed framework's applicability on the RGB data by evaluating it on the Microsoft COCO dataset \cite{coco}. Since the experiments on COCO dataset \cite{coco} do not relate to our proposed study, we report them in the supplementary material of this paper.}

\noindent \textbf{C. Evaluation Metrics:}
\noindent The proposed framework has been evaluated using the \RV{pixel-level recall, precision,} intersection-over-union (IoU), dice coefficient (DC), ROC curves, box-level and mask-level mean average precision ($\mu_{ap}$) computed using IoU$\geq$ 0.5 ($\mu_{ap}^{b:50}$ and $\mu_{ap}^{m:50}$), IoU$\geq$ 0.75 ($\mu_{ap}^{b:75}$ and $\mu_{ap}^{m:75}$), and IoU = $0.5:0.05:0.95$ ($\mu_{ap}^{b}$ and $\mu_{ap}^{m}$), respectively. 

\begin{table}[t]
\centering
\caption{{\small \T{Evaluation of the different segmentation models on the SIXray (S) \cite{miao2019sixray}, GDXray (G) \cite{mery2015gdxray} and Combined (C) dataset. Bold indicates the best performance.}}}
\begin{tabular}{ccccccc}
\toprule
 \multirow{2}{2.6em}{Model} & \multicolumn{3}{c}{IoU} & \multicolumn{3}{c}{DC} \\ \cline{2-7}
       & S & G & C & S & G & C\\ \hline
CIE-Net & \textbf{0.6883} & 0.7723 & \textbf{0.5861} & \textbf{0.8153} & 0.8715 & \textbf{0.7390}\\

\color{black}{CIE-R-Net} & \color{black}{0.6702} & \color{black}{\textbf{0.7852}} & \color{black}{0.5749} & \color{black}{0.8025} & \color{black}{\textbf{0.8796}} & \color{black}{0.7300}\\

PSPNet & 0.6641 & 0.7694 & 0.5728 & 0.7981 & 0.8696 & 0.7283\\ 
SegNet & 0.6559 & 0.7463 & 0.5640 & 0.7921 & 0.8547 & 0.7212\\ 
U-Net & 0.6434 & 0.7384 & 0.5514 & 0.7830 & 0.8495 & 0.7108\\ 
FCN-8 & 0.5792 & 0.6431 & 0.4527 & 0.6973 & 0.7827 & 0.6232\\ 
FCN-32 & 0.5084 & 0.6246 & 0.3931 & 0.6740 & 0.7689 & 0.5643\\
\bottomrule
\end{tabular}
\label{tab:segModelsComp}
\end{table}

\section{Results} \label{sec:results}
\noindent This section reports a thorough evaluation of the proposed framework for extracting and recognizing the contraband items. The purpose of these experiments is two-fold: 1) comparing the performance of our instance segmentation model (CIE-Net) with other state-of-the-art models \cite{msrcnn, maskrcnn, htc, yolact}, and 2) comparing the overall performance of our framework for baggage threat detection with other competitive systems \cite{miao2019sixray, hassan2019, Hassan2020ACCV, hassan2020Sensors}.  \T{At first, we conducted an ablative analysis to assess the performance of different state-of-the-art encoder-decoder and fully convolutional models in our framework. 
We also conduct empirical experimentation to study the effect of the temperature constant ($\tau$) and the effect of utilizing different knowledge distillation loss functions for incremental instance segmentation.} Then, we present the detailed evaluation results of the proposed framework on both GDXray and SIXray datasets in Section \ref{sec:resultsGDXray}-B and Section \ref{sec:resultsSIXray}-C, respectively. Afterward, we report, in Section \ref{sec:combined}-D, the experimentation conducted on the combined datasets.

\noindent \textbf{A. Ablation Study:} \label{sec:ablation}
\T{\noindent We conducted an ablation study to investigate: 1) The optimal choice of the segmentation model; 2) The effect of the temperature constant ($\tau$); 3) The effects of employing different knowledge distillation loss functions in the incremental instance segmentation. \RV{Apart from this, we also conducted rigorous ablation experiments to evaluate the parametric effects of the CIE-Net and its custom CPB, IB, and HB blocks. Due to space constraints, these parametric evaluations are reported within the supplementary material of the article.}
} 

\noindent \textbf{\textit{1. Choice of Segmentation Model:}} \label{sec:ablationA}
In this study, we compared the performance of several state-of-the-art semantic segmentation models, including 
PSPNet \cite{zhao2017pyramid}, SegNet \cite{segnet}, U-Net \cite{ronneberger2015unet}, FCN-8 and FCN-32  \cite{fcn8} with our proposed CIE-Net model \T{for the extraction of isolated and \RV{overlapping} contraband items and their instances depicted within the grayscale and colored baggage X-ray scans. We further want to notify that to fairly compare all the models, we have trained them incrementally using the proposed $L_s$ loss function where each model, including the CIE-Net model, was implemented using ResNet\textsubscript{101} \cite{he2016deep}.  We dubbed this CIE-Net variant as CIE-R-Net to differentiate it from the CIE-Net build with our custom backbone. 

\noindent The comparison results are reported in Table \ref{tab:segModelsComp}, where we can see that the proposed CIE-Net produced the best performance in terms of both IoU and DC metrics for the SIXray \cite{miao2019sixray}, GDXray \cite{mery2015gdxray}, and the combined datasets.}

\begin{figure}[htb]  
\includegraphics[width=1\linewidth]{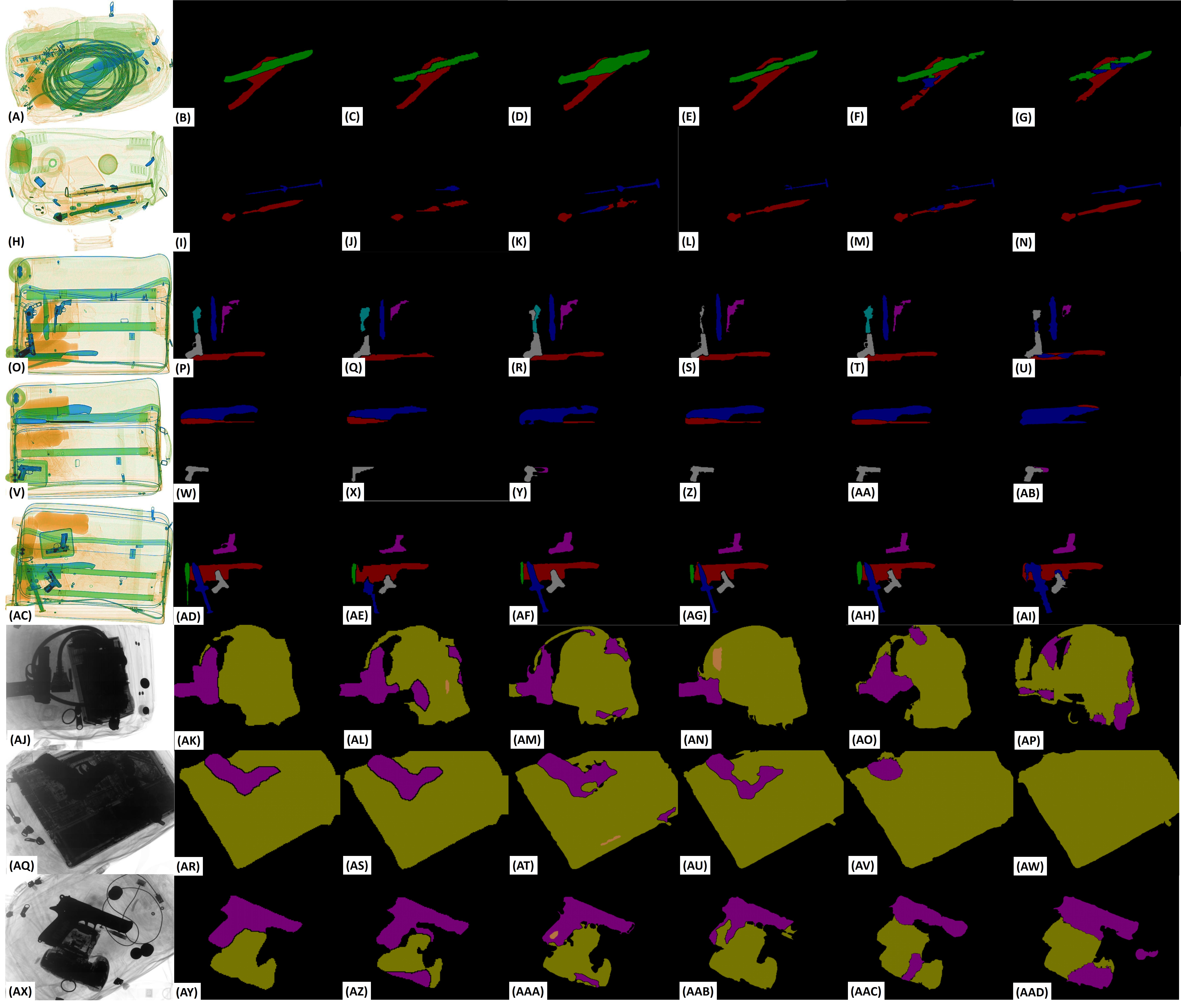}
\caption{ \small \color{black}{Extraction of contraband items (and their instances) using different segmentation models. From left: Original X-ray scan, CIE-Net, PSPNet \cite{zhao2017pyramid}, SegNet \cite{segnet}, U-Net \cite{ronneberger2015unet}, FCN-8, and FCN-32 \cite{fcn8}. Zoom-in for better visualization.} }
\centering
\label{fig:visualComparisonAblative}
\end{figure}

\noindent Moreover, Figure \ref{fig:visualComparisonAblative} depicts a qualitative comparison showing segmentation results on samples from the SIXray and GDXray dataset. We can observe here that the CIE-Net produces better extraction results, especially for the examples in Figure \ref{fig:visualComparisonAblative} (A), (AJ), (AQ) and (AX). \RV{This better performance emanates from integrating the CPB, IB, and HB blocks in our model as showcased through rigorous parametric evaluations discussed in the supplementary material.} Also, such synergy allows better extraction of contraband items by retaining global contextual information about the contraband items, even at the sparsest level of decomposition, while integrating finer features from the consecutive encoder part through the skip-connections.

\noindent \textbf{\textit{2. Effects of the Temperature Parameter:}}
In this experiment, we varied $\tau$ from 0.1 to 3 and measured its effects on the segmentation performance for GDXray,  SIXray, and the combined datasets. The results, depicted in Table \ref{tab:effectTemp}, indicate $\tau=2$ and $\tau=1.5$ as the best values for the GDXray and the SIXray datasets, respectively. $\tau=2$ also yields the highest performance on the combined dataset.  These results suggested framing the optimal values of $\tau$ within the range $[1.5, 2]$.
\begin{table}[t]
\center
\caption{  {\small \color{black}{Effects of varying the temperature paramter $\tau$ (in terms of IoU).}}}
\begin{tabular}{cccc}
\toprule
$\tau$ & GDXray & SIXray & Combined \\ \hline
0.1 & 0.4462 &  0.4106 & 0.2614\\
0.2 & 0.5013  & 0.4731 & 0.3053\\
0.5 & 0.6425 & 0.5632 & 0.3987\\
1 & 0.7341 & 0.6482 & 0.5014\\
1.5 & 0.7524 & \textbf{0.6883} & 0.5659\\
2 & \textbf{0.7723} & 0.6697 & \textbf{0.5861}\\
2.5 & 0.7214 & 0.6021 & 0.5543\\
3 & 0.6642 & 0.5364 & 0.4471\\
\bottomrule
\end{tabular}
\label{tab:effectTemp}
\end{table}


\noindent \textbf{\textit{3. Knowledge Distillation Loss Function:}} \label{sec:ablationC}
This objective of this ablation study is to compare $L_{mi}$ function with  other state-of-the-art knowledge distillation loss functions, such as Output Distillation Loss ($L_{od}$) \cite{kdil}, Modified Deep Model Consolidation \cite{dmc} Loss ($L_{ds}$) (proposed in \cite{kdils}), Similarity-Preserving Knowledge Distillation Loss ($L_{sp}$) \cite{spkd}, and Joint Classification and Distillation Loss ($L_{cd}$) \cite{icarl}, in our  framework.   The comparison was made by switching the $L_{mi}$ term in Eq. \ref{equ:loss} with these distillation loss functions. 

\noindent In what comes next, we denote by $A_{k-1}$ and $A_{k}$, the models trained in the previous iteration (from 1 to $k-1$), and in the current iteration $k$, respectively,  $N_{t_o}$ denotes the total number of training examples belonging to the previously learned classes,  $X^{o}_i$, $i=1:N_{t_o}$, denotes an old  training sample,
$C_{W_o}$ denotes the total number of old classes, and $F$ represents the Frobenius norm.
\noindent Moreover, $L_{od}$ minimizes the cross-entropy loss between the prediction  of $A_{k-1}$ and  $A_k$ and is  expressed below:
\begin{equation}
  L_{od} =  \frac{1}{N_{t_o}}\sum_{i=0}^{N_{t_o} -1}\sum_{j=0}^{C_{W_o}-1} (p(l_{i,j}^{o,\tau})_{A_{k-1}})\log(p(l_{i,j}^{o,\tau})_{A_{k}}),
\label{eq:eq9}
\end{equation}

\noindent $L_{ds}$ minimizes the disparities between the latent space feature representation of $A_{k-1}$ and  $A_k$ and defined as:
\begin{equation}
  L_{ds} =  \frac{1}{N_{t_o}}\sum_{i=0}^{N_{t_o} -1} ||\mathcal{E}_{k-1}(X^{o}_i)-\mathcal{E}_k(X^{o}_i)||^2_F,
\label{eq:eq10}
\end{equation}
\noindent where $\mathcal{E}_{k-1}$ and $\mathcal{E}_{k}$ are the latent space vectors related to $A_{k-1}$ and $A_{k-1}$, respectively. 

\noindent $L_{sp}$  minimizes the disparities between the activation similarity matrices ($\mathcal{S}$) \cite{spkd}, and  expressed as:
\begin{equation}
  L_{sp} =  \frac{1}{N_{t_o}}\sum_{i=0}^{N_{t_o} -1} ||\mathcal{S}_{k-1}(X^{o}_i)-\mathcal{S}_k(X^{o}_i)||^2_F.
\label{eq:eq11}
\end{equation}

\noindent The joint classification and distillation loss $L_{cd}$, proposed in iCaRL \cite{icarl}, is expressed as follows:
\begin{equation}
L_{cd} = \mathcal{L}_{CE}(t_{i,j}^{n}, l_{i,j}^{n,\tau}) + \mathcal{L}_{CE}(t_{i,j}^{o}, l_{i,j}^{o,\tau}),
\label{eq:eq12}
\end{equation}
where $\mathcal{L}_{CE}$ is the standard cross-entropy loss function and the other terms are as previously defined in Eq. (\ref{eq:Lo}) and (\ref{eq:Ln}). Note that unlike the previous knowledge distillation loss functions, which are plugged as a replacement to $L_{mi}$, $L_{cd}$  is used as a replacement of $L_s$ in Eq. \ref{equ:loss}. This is because  $L_{cd}$ minimizes both the loss for learning new class representations and the distillation loss for retaining the previously learned classes. 

\noindent The comparison of the loss functions is reported in Table \ref{tab:loss} in term of IoU score where we can see that the proposed framework achieves 2.83\%, 2.16\%, and 6.50\% performance improvements over the second-best $L_{sp}$ \cite{spkd} on GDXray \cite{mery2015gdxray}, SIXray \cite{miao2019sixray}, and the combined dataset, respectively. These improvements emanate because of the synergy between $L_n$, $L_o$, and $L_{mi}$ that not only retains the prior knowledge while learning new classes but also enables the network to analyze the mutual relationships between the knowledge representations of the old and the new instances via Bayesian inference, unlike its competitors, that mostly rely on the spatial \cite{kdil} and contextual \cite{spkd} differences between knowledge representations.

\begin{table}[t]
    \centering
    \caption{\color{black}{Comparison of $L_{mi}$ with state-of-the-art knowledge distillation loss functions in terms of IoU. \RV{To ensure fairness, we used CIE-Net with all the loss functions.}} }

    \begin{tabular}{cccc}
        \toprule
        \color{black}{Loss Functions} &  \color{black}{GDXray \cite{mery2015gdxray}} & \color{black}{SIXray \cite{miao2019sixray}} & \color{black}{Combined}\\
        \hline
        \color{black}{$L_{mi}$} & \color{black}{\textbf{0.7723}} & \color{black}{\textbf{0.6883}} & \color{black}{\textbf{0.5861}} \\
        \color{black}{$L_{sp}$ \cite{spkd}} & \color{black}{0.7504} & \color{black}{0.6734} & \color{black}{0.5480} \\
        \color{black}{$L_{od}$ \cite{kdil}} & \color{black}{0.7349} & \color{black}{0.6162} & \color{black}{0.5018} \\
        \color{black}{$L_{ds}$ \cite{kdils}} & \color{black}{0.7421} & \color{black}{0.6395} & \color{black}{0.5237} \\
        \color{black}{$L_{cd}$ \cite{icarl}} & \color{black}{0.6052} & \color{black}{0.4793} & \color{black}{0.2746} \\
        \bottomrule
    \end{tabular}
    \label{tab:loss}
\end{table}

\begin{table}[t]
    \centering
    \caption{ \RV{Comparison of the proposed framework with state-of-the-art solutions for extracting baggage threats. Bold indicates the best performance, while the second-best scores are underlined.}}
    \begin{tabular}{ccccc}
        \toprule
        Metric & Method &  GDXray & SIXray & Combined \\
        
        \hline
        IoU &Proposed & \textbf{0.7723} & \textbf{0.6883} & \textbf{0.5861} \\
        &MS RCNN \cite{msrcnn} & 0.7201 & 0.6484 & 0.5482 \\
        &Mask RCNN \cite{maskrcnn} & 0.7098 & 0.6381 & 0.5243 \\
        &HTC \cite{htc} & 0.7364 & \underline{0.6559} & \underline{0.5804} \\
        &YOLACT \cite{yolact} & 0.7089 & 0.6110 & 0.4937 \\
        &\color{black}{DSRL \cite{wang_cvpr2020}} & \color{black}{\underline{0.7421}} & \color{black}{0.6542} & \color{black}{0.5709}   \\
        &\color{black}{MvRF-CNN \cite{akilan2020}} & \color{black}{0.6982} & \color{black}{0.6016} & \color{black}{0.4918}  \\
        &\color{black}{TST-$L_s$ \cite{Hassan2020ACCV}} & \color{black}{0.6851} & \color{black}{0.5874} & \color{black}{0.4285} \\\hline
        
        \color{black}{DC} & \color{black}{Proposed} & \color{black}{\textbf{0.8715}} & \color{black}{\textbf{0.8153}} & \color{black}{\textbf{0.7390}} \\
        &\color{black}{MS RCNN \cite{msrcnn}} & \color{black}{0.8372} & \color{black}{0.7867} & \color{black}{0.7081}\\
        &\color{black}{Mask RCNN \cite{maskrcnn}} & \color{black}{0.8302} & \color{black}{0.7790} & \color{black}{0.6879} \\
        &\color{black}{HTC \cite{htc}} & \color{black}{0.8481} & \color{black}{\underline{0.7921}} & \color{black}{\underline{0.7344}} \\
        &\color{black}{YOLACT \cite{yolact}} & \color{black}{0.8296} & \color{black}{0.7585} & \color{black}{0.6610} \\
        &\color{black}{DSRL \cite{wang_cvpr2020}} & \color{black}{\underline{0.8519}} & \color{black}{0.7909} & \color{black}{0.7268}  \\
        &\color{black}{MvRF-CNN \cite{akilan2020}} & \color{black}{0.8222} & \color{black}{0.7512} & \color{black}{0.6593} \\
        &\color{black}{TST-$L_s$ \cite{Hassan2020ACCV}} & \color{black}{0.8131} & \color{black}{0.7400} & \color{black}{0.5999} \\\hline
        
        \color{black}{Recall} &\color{black}{Proposed} & \color{black}{\textbf{0.8643}} & \color{black}{\textbf{0.8057}} & \color{black}{\textbf{0.7391}} \\
        &\color{black}{MS RCNN \cite{msrcnn}} & \color{black}{0.8238} & \color{black}{0.7613} & \color{black}{0.6846} \\
        &\color{black}{Mask RCNN \cite{maskrcnn}} & \color{black}{0.8183} & \color{black}{0.7542} & \color{black}{0.6653} \\
        &\color{black}{HTC \cite{htc}} & \color{black}{0.8392} & \color{black}{\underline{0.7736}} & \color{black}{\underline{0.7284}} \\
        &\color{black}{YOLACT \cite{yolact}} & \color{black}{0.8195} & \color{black}{0.7461} & \color{black}{0.6548} \\
        &\color{black}{DSRL \cite{wang_cvpr2020}} & \color{black}{\underline{0.8407}} & \color{black}{0.7705} & \color{black}{0.7173}  \\
        &\color{black}{MvRF-CNN \cite{akilan2020}} & \color{black}{0.8196} & \color{black}{0.7344} & \color{black}{0.6419}  \\
        &\color{black}{TST-$L_s$ \cite{Hassan2020ACCV}} & \color{black}{0.8092} & \color{black}{0.7269} & \color{black}{0.5764} \\\hline
        
        \color{black}{Precision} &\color{black}{Proposed} & \color{black}{\textbf{0.8952}} & \color{black}{\textbf{0.8348}} & \color{black}{\textbf{0.7401}}\\
        &\color{black}{MS RCNN \cite{msrcnn}} & \color{black}{0.8564} & \color{black}{0.8153} & \color{black}{0.7269} \\
        &\color{black}{Mask RCNN \cite{maskrcnn}} & \color{black}{0.8439} & \color{black}{0.8072} & \color{black}{0.7154} \\
        &\color{black}{HTC \cite{htc}} & \color{black}{0.8607} & \color{black}{\underline{0.8236}} & \color{black}{\underline{0.7318}} \\
        &\color{black}{YOLACT \cite{yolact}} & \color{black}{0.8353} & \color{black}{0.7669} & \color{black}{0.6703} \\
        &\color{black}{DSRL \cite{wang_cvpr2020}} & \color{black}{\underline{0.8736}} & \color{black}{0.8125} & \color{black}{0.7311} \\
        &\color{black}{MvRF-CNN \cite{akilan2020}} & \color{black}{0.8245} & \color{black}{0.7801} & \color{black}{0.6786} \\
        &\color{black}{TST-$L_s$ \cite{Hassan2020ACCV}} & \color{black}{0.8173} & \color{black}{0.7614} & \color{black}{0.6256} \\
        \bottomrule
    \end{tabular}
    \label{tab:iou}
\end{table}
\begin{figure}[htb]
    \centering
    \includegraphics[width=1\linewidth]{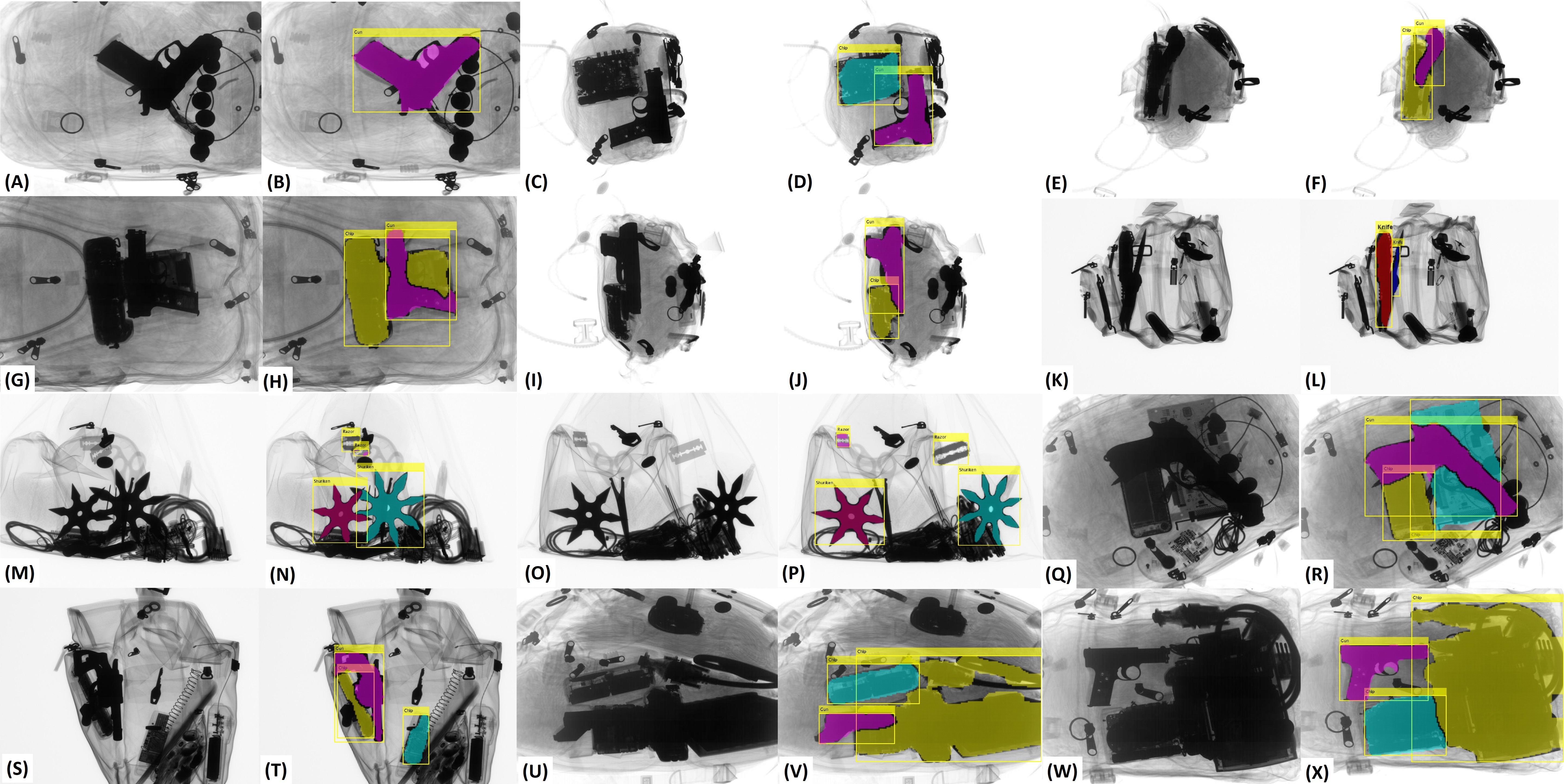}
    \caption{ {\small \color{black}{GDXray \cite{mery2015gdxray}: Examples of occluded and overlapping items detection. Please zoom-in for better visualization.}
    }} 
    \label{fig:visualGDXray}
\end{figure}

\begin{figure*}[t]  
\centering
\includegraphics[scale=0.215]{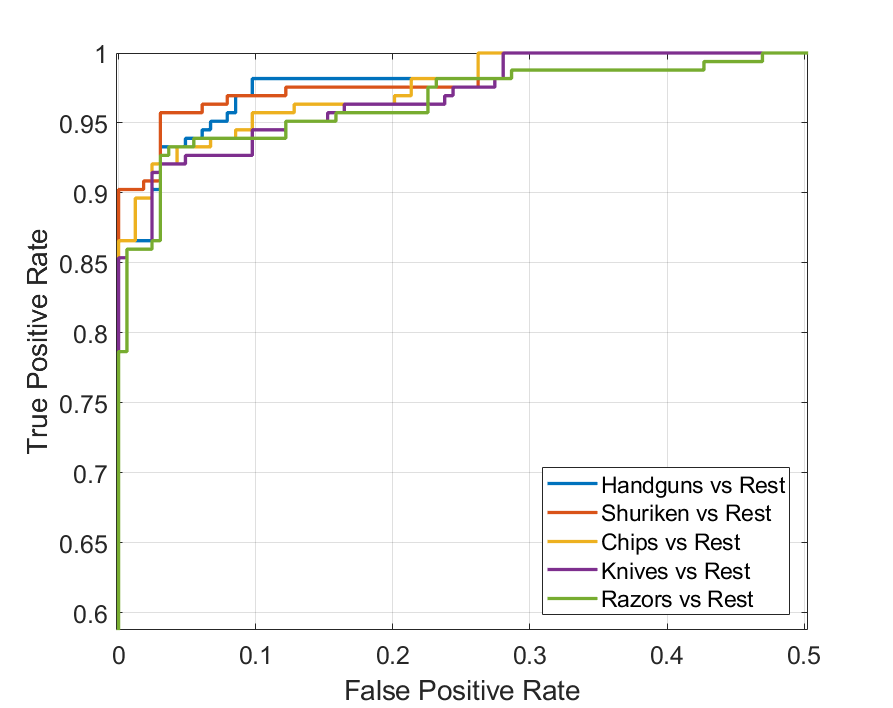}
\includegraphics[scale=0.255]{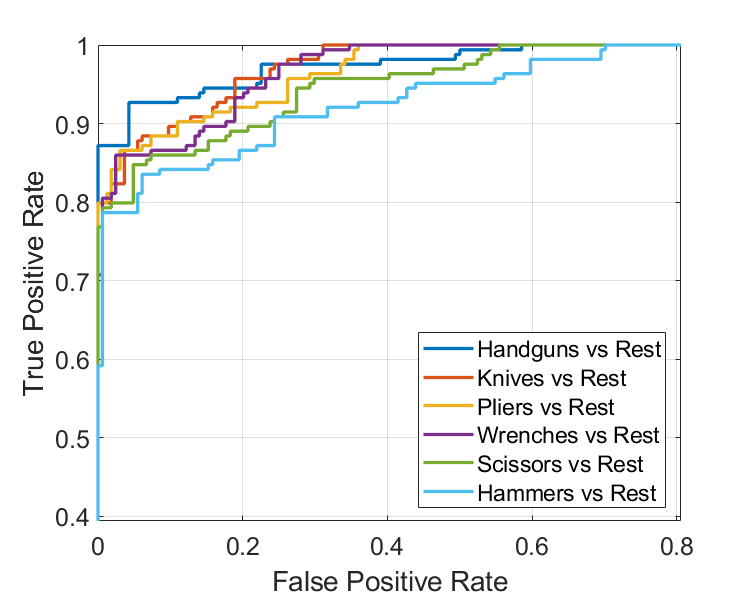}
\includegraphics[scale=0.215]{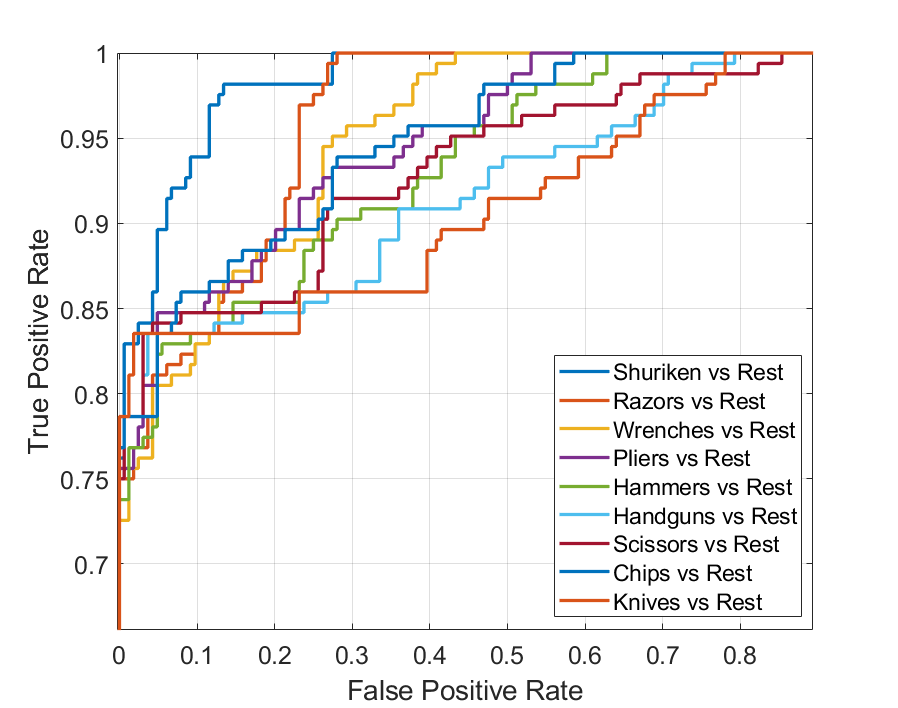}
\centerline {(a) \hspace{4.6cm} (b) \hspace{4.6cm} (c) }
\caption{ \small \T{Performance evaluation of CIE-Net in terms of ROC for extracting contraband items from (a) GDXray dataset, (b) SIXray dataset, and (c) the combined dataset.}}
\label{fig:ROC}
\end{figure*}


\noindent \textbf{B. Evaluations on GDXray Dataset:} \label{sec:resultsGDXray}
\noindent The CIE-Net was trained for two iterations on GDXray as this dataset contains at most two overlapping instances of the same contraband item. Table \ref{tab:gd1} shows the performance comparison  against the state-of-the-art schemes.  We can observe that our framework achieves 4.08\%  and 28.39\% 
better performance than the second-best HTC \cite{htc} and the YOLACT \cite{yolact}, respectively, in terms of $\mu_{ap}^{m}$. 
Furthermore, it outperforms the second-best performing HTC \cite{htc} by 2.13\% in terms of $\mu_{ap}^{b}$. However, for $\mu_{ap}^{b:50}$, the best performance is achieved by \RV{the original TST \cite{Hassan2020ACCV} (dubbed TST$_o$) from which the proposed framework lags by 11.53\%. However, this is an unfair comparison since TST \cite{Hassan2020ACCV} is trained conventionally using the large-scale well-annotated training data. In contrast, the proposed framework is trained incrementally on small-scale training batches. Moreover, under fair comparison with the incremental TST \cite{Hassan2020ACCV} scheme, dubbed TST-$L_s$, the proposed CIE-Net is leading by 3.63\%.}
\begin{table}[t]
\center
\caption{ \RV{Comparison of the proposed framework with state-of-the-art solutions for extracting contraband items. Bold indicates the best scores, while ‘-’ means that the metric is not computed.}}
\begin{tabular}{llllllll}
\toprule
D & M              & $\mu_{ap}^m$ & $\mu_{ap}^{m:50}$ & $\mu_{ap}^{m:75}$ & $\mu_{ap}^{b}$ & $\mu_{ap}^{b:50}$ & $\mu_{ap}^{m:75}$ \\ \toprule
G & PF & \textbf{0.5068} & \textbf{0.7902} & \textbf{0.5006} & \textbf{0.6101} & 0.8556 & \textbf{0.6462}\\

& MSR & 0.4584 & 0.7283 & 0.4986 & 0.5564	& 0.8091 & 0.6033\\

& MR & 0.4311 & 0.7194	& 0.4893 & 0.5282 & 0.7833 & 0.5842 \\

& HTC & 0.4861 & 0.7706 & 0.4997 & 0.5971 & 0.8314 & 0.6324 \\

& YT & 0.3629 & 0.6518 & 0.3794 & 0.4852 & 0.7478 & 0.5491\\ 

& CST$_o$\RV{*} & - & - & - & - & 0.9343 & - \\

& TST$_o$\RV{*} & - & - & - & - & \textbf{0.9672} & - \\

& \RV{TST$_i$} & \RV{-} & \RV{-} & \RV{-} & \RV{-} & \RV{0.8245} & \RV{-} \\

& \RV{CST$_i$} & \RV{-} & \RV{-} & \RV{-} & \RV{-} & \RV{0.8169} & \RV{-} \\
& TSD\RV{*} & - & - & - & - & 0.9162 & - \\ \hline

S & PF & \textbf{0.4795} & \textbf{0.6893} & \textbf{0.4872} & \textbf{0.5367} & 0.7653 & \textbf{0.5374} \\

& MSR & 0.4017 & 0.6347 & 0.4063 & 0.4653 & 0.6756 & 0.4782\\

& MR & 0.3654 & 0.5973 & 0.3592 & 0.4182 & 0.6326 & 0.4067 \\

& HTC & 0.4525 & 0.6629 & 0.4538 & 0.5082 & 0.7384 & 0.5021 \\

& YT & 0.3355 & 0.5632 & 0.3190 & 0.3811 & 0.6237 & 0.3643\\ 

& CST$_o$\RV{*} & - & - & - & - & \textbf{0.9595} & - \\

& TST$_o$\RV{*} & - & - & - & - & 0.9516 & - \\

& \RV{TST$_i$} & \RV{-} & \RV{-} & \RV{-} & \RV{-} & \RV{0.7248} & \RV{-} \\

& \RV{CST$_i$} & \RV{-} & \RV{-} & \RV{-} & \RV{-} & \RV{0.7351} & \RV{-} \\

& TSD\RV{*} & - & - & - & - & 0.6457 & - \\

& CHR & - & - & - & - & 0.5760 & - \\ \hline

C & PF & \textbf{0.4059} & \textbf{0.6249} & \textbf{0.4153} & \textbf{0.4862} & \textbf{0.7249} & \textbf{0.4983} \\

& MSR & 0.3591 & 0.5986 & 0.3865 & 0.4023 & 0.6298 & 0.4572 \\

& MR & 0.3129 & 0.5542 & 0.3301 & 0.3627 & 0.5983 & 0.3821 \\

& HTC & 0.4023 & 0.6173 & 0.4102 & 0.4752 & 0.7203 & 0.4859\\

& YT & 0.3098 & 0.5286 & 0.3123 & 0.3561 & 0.5937 & 0.3696\\ 

& \RV{TST$_i$} & \RV{-} & \RV{-} & \RV{-} & \RV{-} & \RV{0.6718} & \RV{-} \\

& \RV{CST$_i$} & \RV{-} & \RV{-} & \RV{-} & \RV{-} & \RV{0.6526} & \RV{-} \\


\bottomrule
\end{tabular}
 \begin{tablenotes}[flushleft]
\RV{Abbreviations: D: Dataset, G: GDXray \cite{mery2015gdxray}, S: SIXray \cite{miao2019sixray}, C: Combined Dataset, M: Methods, 
PF: Proposed Framework, MSR: Mask Scoring R-CNN \cite{msrcnn}, MR: Mask R-CNN \cite{maskrcnn}, and YT: YOLACT \cite{yolact}. Moreover, '*' indicates unfair comparison.}
    \end{tablenotes}
    
\label{tab:gd1}
\end{table}
Apart from this, the CIE-Net performance is further evaluated through the ROC curves, as shown in Figure \ref{fig:ROC} (a). These curves are generated considering the pixel-level recognition, i.e., the pixel for each item (along with their instances) are treated as one and the rest of the pixels as zero (a typical binary classification). We can observe that the instance-aware CIE-Net achieved the minimum AUC score of 0.9818 for extracting \textit{razors}. Due to space constraints, we report the detailed AUC score for each item (for all the datasets) within the source code repository\textsuperscript{\ref{note1}}. 

\T{\noindent Moreover, we also compared the performance of the proposed CIE-Net against the state-of-the-art semantic and instance segmentation frameworks. The results are reported in Table \ref{tab:iou}, where we can see that on GDXray, in terms of IoU, CIE-Net achieves 3.91\% improvements over the DSRL \cite{wang_cvpr2020} framework. Similarly, it outperforms HTC \cite{htc} by 4.64\%. 

\noindent In addition to this, we fairly compared the proposed framework with TST \cite{Hassan2020ACCV} by incrementally training it using the same experimental protocols and the proposed $L_s$ function, where the proposed framework achieves 11.29\% superior results, in terms of IoU, as evident from Table \ref{tab:iou}. The degradation in the TST's performance stems from the fact that during incremental training, it is more susceptible to forgetting the prior learned categories while adapting to new class representations since it employs a contour-driven strategy towards recognizing contraband items \cite{Hassan2020ACCV}}.

\noindent Moreover, the performance of CIE-Net on the GDXray dataset is further analyzed through visual examples, as shown in Figure \ref{fig:visualGDXray}.
The GDXray contains at most two overlapping instances of the same items, e.g., see Figure \ref{fig:visualGDXray} (N, L, P, R, V, and X).
Here, we can appreciate the extraction performance of CIE-Net by observing two extracted occluded \textit{knives} in (L) and occluded \textit{shuriken} in (N, P). We can also observe how accurately the low-intensity \textit{razors} have been segmented by the CIE-Net in Figure \ref{fig:visualGDXray} (N, P).

\noindent \textbf{C. Evaluations on SIXray Dataset:} \label{sec:resultsSIXray}
\noindent For the SIXray dataset, the training was conducted for six iterations since there are at most six instances of the same item in this dataset. Table \ref{tab:gd1} shows the model's comparison against the state-of-the-art instance segmentation algorithms. CIE-Net achieves 5.63\% improvements in terms of $\mu_{ap}^{m}$ against the second-best HTC \cite{htc} and 30.03\% higher than the least good performing YOLACT \cite{yolact}. It also achieves 5.31\% superior results than the existing solutions in terms of $\mu_{ap}^{b}$.
\RV{For $\mu_{ap}^{b:50}$, the CIE-Net comes third after the original CST \cite{hassan2019} (dubbed CST$_o$) and the original TST \cite{Hassan2020ACCV} (dubbed TST$_o$) scheme. However, this comparison is unfair, and the increased performance of CST$_o$ \cite{hassan2019} and TST$_o$ \cite{Hassan2020ACCV} here emanates from the conventional fine-tuning strategy, which utilizes the whole training dataset. Under fair comparison with incremental TST \cite{Hassan2020ACCV} (dubbed TST$_i$) and CST (dubbed CST$_i$), the CIE-Net is leading by 5.29\% and 3.94\%, respectively.} 
Apart from this, the CIE-Net performance on SIXray is further evaluated through the ROC curves shown in Figure \ref{fig:ROC} (b). Here, we can observe that the proposed framework achieves the best AUC score for extracting the \textit{handguns}. 
\T{In addition to this, the segmentation performance of our framework can be analyzed through the mean IoU score in Table \ref{tab:iou}, showing the best score of 0.6883, leading the second-best HTC \cite{htc} by 4.70\%.
}

\begin{figure}[t]  
\includegraphics[width=1\linewidth]{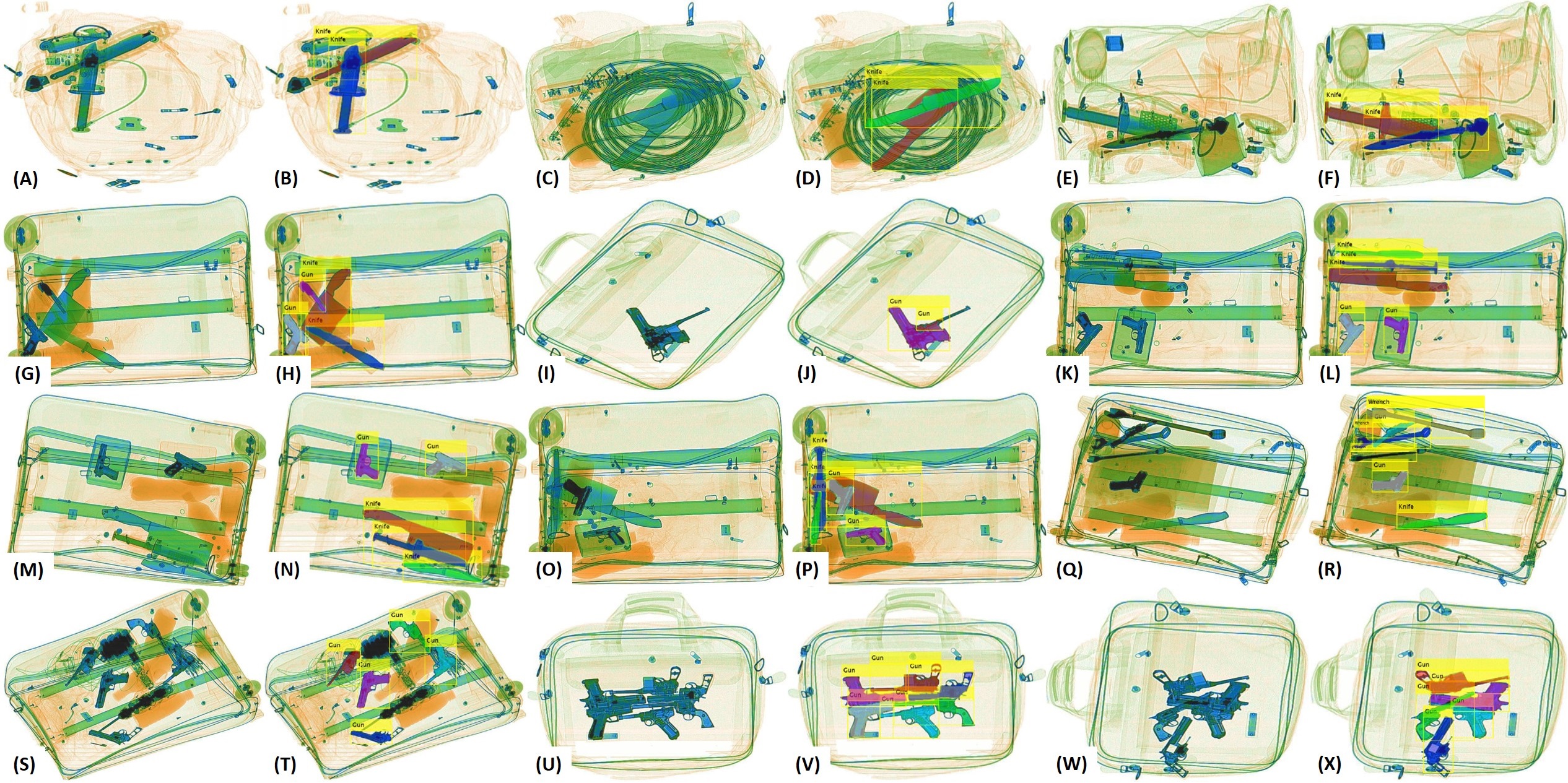}
\caption{ {\small \color{black}{SIXray \cite{miao2019sixray}: Examples of occluded and overlapping objects detection. Please zoom-in for better visualization.}}.
}
\centering
\label{fig:sixray-qual1}
\end{figure}

\noindent In Figure \ref{fig:sixray-qual1}, we report the qualitative evaluation showcasing examples of successfully extracted overlapping items, e.g., two items (B, D, F, H) and three items (N, P, R) and up to six items (V, X). In these examples,  we can appreciate the potential of the instance-aware CIE-Net in accurately recognizing the extremely merged items, e.g., an instance of \textit{guns} in Figure \ref{fig:sixray-qual1} (J, V, and X). 

\noindent \textbf{D. Evaluations on Combined Dataset:} \label{sec:combined}
\noindent 
We also evaluated the proposed framework on the combined dataset. The results on the combined dataset are reported in Table \ref{tab:iou} and \ref{tab:gd1}. \T{From Table \ref{tab:gd1}, we can observe that CIE-Net achieved the best $\mu_{ap}^{b:50}$ performance of 0.7249, outperforming the second-best framework by 0.6345\%.} Furthermore, we can also notice the performance gain of 23.67\%  over YOLACT \cite{yolact} in terms of $\mu_{ap}^{m}$. 
\RV{Moreover, in terms of recall and precision, the CIE-Net is outperforming the second-best framework by 1.44\%, and 1.12\%, respectively (see Table \ref{tab:iou}). 
}

\noindent In addition to this, Figure \ref{fig:ROC} (c) further depicts the ROC performance of instance-aware CIE-Net for extracting contraband items. Here, we can see that the minimum score is obtained for \textit{knives} and \textit{handguns} (i.e., AUC of 0.9133 and 0.9212, respectively).
\begin{figure}[htb]  
\label{fig:sixrayqual}
\includegraphics[width=1\linewidth]{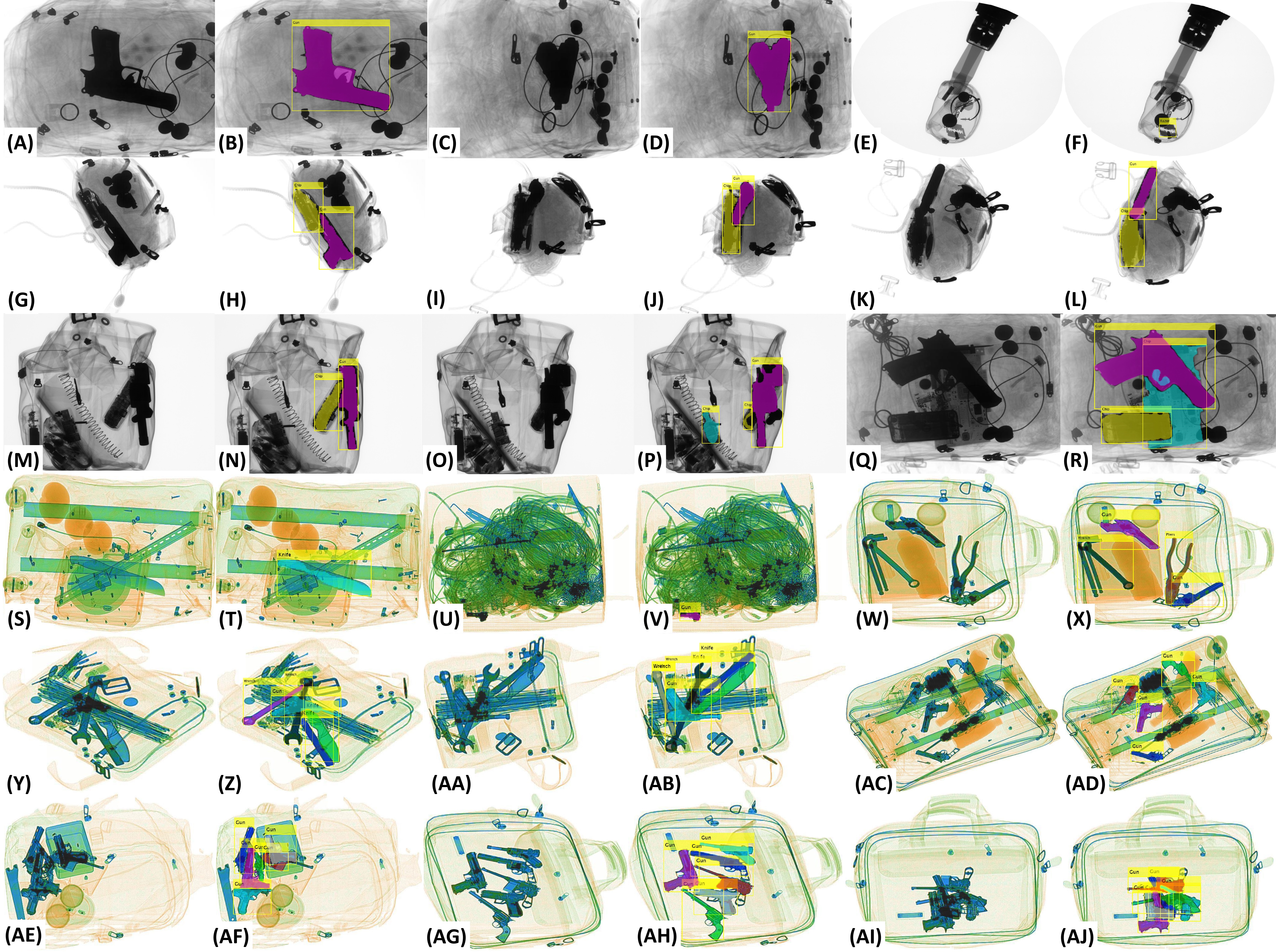}
\caption{ {\small \color{black}{Examples of occluded and overlapping objects detection on combined dataset. Zoom-in for better visualization.} 
}}
\centering
\label{fig:combined-qual}
\end{figure}

\noindent Figure \ref{fig:combined-qual} showcases some qualitative examples derived from the combined dataset, which illustrates the capacity of CIE-Net for extracting instances of overlapped items despite the large differences of the scan properties in GDXray and SIXray datasets.
In Figure \ref{fig:combined-qual} (F), we can observe  how effectively the \textit{razor} is extracted in such a cluttered scenario.
Figure \ref{fig:combined-qual} (N, R) depicts examples whereby our framework  robustly differentiated between  merged \textit{gun} and \textit{chip} instances. Figure \ref{fig:combined-qual} (T) depicts a reasonable  extraction of the  occluded \textit{knife}. The performance of the CIE-Net can also be appreciated on more highly challenging scans such as (V), where a \textit{gun} has been extracted from an extremely cluttered environment, (AB) in which two overlapping \textit{wrenches}, two overlapping \textit{knives} and a barely visible \textit{gun} have been recognized, (AF) and (AJ) from which six extremely \RV{overlapping} \textit{guns} are accurately extracted. In Figure \ref{fig:combined-qual} (AF, AJ), in particular,  we can appreciate the capacity of CIE-Net in accurately recognizing six instances of \textit{guns} under extreme occlusion.  

\T{
\noindent \textbf{E. Comparison of Run-time Performance:}
\noindent Apart from evaluating the proposed scheme's detection performance, we also analyzed its run-time performance and compared it with state-of-the-art methods. The comparison is reported in Table \ref{tab:time}. Here, we can see that the proposed CIE-Net lags behind the state-of-the-art frameworks in terms of efficiency. This is due to the design choice of CIE-Net to focus more on accurately extracting the contraband items rather than achieving efficiency. 

\noindent Due to this, the CIE-Net is slower than the other lightweight models like YOLOv3 \cite{yolov3}, and CST \cite{hassan2019}. However, we also want to highlight that the proposed framework is an instance segmentation scheme (unlike region-based YOLOv3 \cite{yolov3} and contour-based CST \cite{hassan2019} detectors), and it gives the best trade-off between contraband items extraction (see Table \ref{tab:gd1}) and run-time performance (see Table \ref{tab:time}).
\begin{table}[htb]
\center
\caption{ {\small \color{black}{Comparison of the run-time performance. The scores here represent the mean inference time of the two datasets.} Bold indicates the best performance while the second-best performance is underlined.} }
\begin{tabular}{cc}
\toprule
Method & Time Performance (sec)\\\hline
YOLOv3 \cite{yolov3} & \textbf{0.023}\\
CST \cite{hassan2019} & \textbf{0.023}\\
RetinaNet \cite{retinanet} & \underline{0.033}\\
YOLCAT  \cite{yolact} & 0.036\\
CIE-Net (Proposed) & 0.072\\
Mask R-CNN \cite{maskrcnn} & 0.141\\
MS R-CNN  \cite{msrcnn} & 0.156\\
HTC \cite{htc} & 0.311\\
\bottomrule
\end{tabular}
\label{tab:time}
\end{table}

\RV{
\noindent \textbf{F. Failure Cases:}
\noindent Although the proposed framework achieves remarkable performance towards extracting \RV{overlapping} contraband items (and their instances), as evident from Table \ref{tab:iou}, and \ref{tab:gd1},  there are some cases where the CIE-Net turns out to be limited, especially on the negative SIXray scans (see pairs (A, B), (C, D), (E, F), (K, L) and (M, N) in Figure \ref{fig:failure}), producing pixel-level false positives and false negatives due to spatial and contextual similarity between the normal and suspicious baggage content within the X-ray scans. False positives are produced when the background regions (within the candidate scan) are misclassified as threatening items by the proposed framework as shown in Figure \ref{fig:failure}-B, D, F, L, and N. Moreover, false negatives are generated when the region of the contraband item is misclassified as background. For example, see the missed portion of \textit{shuriken} in Figure \ref{fig:failure} (X). Apart from this, in some cluttered cases, the proposed CIE-Net produced over-segmentation results by confusing between different instances of the suspicious items (as shown in Figure \ref{fig:failure}-H, P, R, T, V, and Z). Although all these types of failures were seen rarely during the experimentation, they can be remedied through postprocessing schemes such as blob filtering, region-opening, and region-filling schemes. } 

\begin{figure}[htb]  
\includegraphics[width=1\linewidth]{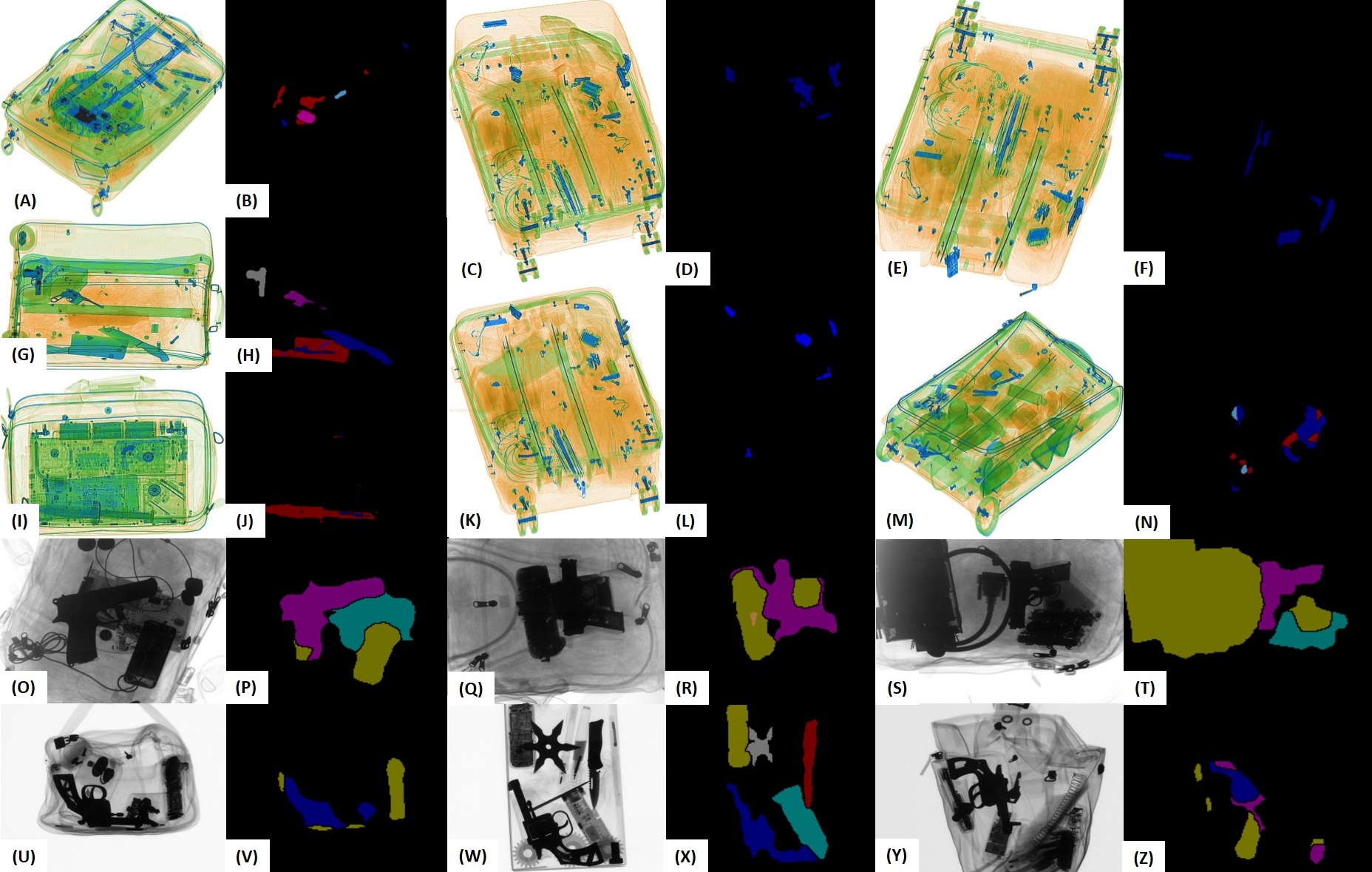}
\caption{ \small \T{Failure cases in GDXray and SIXray datasets. Blue in (B, D, F, H, J, L, N) and red color in (B, H, J, N, R, X) represent \textit{knives}. White in (H), magenta in (B, H, P, R, T, Z), cyan in (B, N), and blue color in (V, X, Z) represent \textit{handguns}.  Yellow and cyan color in (P, R, T, V, X, Z) depicts \textit{chips}. White color in (V) represents the \textit{shuriken}. Zoom-in for better visualization.}}
\centering
\label{fig:failure}
\end{figure}
}

\vspace{-.3cm}
\section{Discussion} \label{sec:discussion}

\noindent An overview of the results in Tables \ref{tab:iou} and  \ref{tab:gd1} convey that the proposed  CIE-Net, employed within the incremental instance segmentation framework, shows neat performance improvement over standard models such as Mask Scoring R-CNN \cite{msrcnn}, Mask R-CNN \cite{maskrcnn}, Hybrid Task Cascade \cite{htc} and YOLACT \cite{yolact}). \T{It also exhibits a competitive performance with models specifically designed for extracting threatening items from X-ray scans (such as CST \cite{hassan2019}, TST \cite{Hassan2020ACCV}, and  TSD\cite{hassan2020Sensors}).}
 
\T{\noindent The CIE-Net lags from the fine-tuning-based contour instance segmentation framework TST \cite{Hassan2020ACCV} in terms of $\mu_{ap}^{b:50}$. However, over the incremental TST-${L_s}$ \cite{Hassan2020ACCV} version, it achieves  11.29\% on the GDXray dataset, 14.65\% improvements on the SIXray dataset, and 26.88\% on the combined dataset in terms of IoU (see Table \ref{tab:iou}). The TST \cite{Hassan2020ACCV} possesses the capacity to eliminate unwanted baggage contours due to extensive fine-tuning on the large-scale training datasets, resulting in the better extraction of the threatening items. In return, the  TST \cite{Hassan2020ACCV} requires large-scale well-annotated training data to achieve optimal performance. Indeed, when we trained TST \cite{Hassan2020ACCV} framework incrementally on small-scale training batches using the proposed $L_s$ loss function to compare it with the CIE-Net fairly, it produces degraded performance, as evidenced from the results mentioned above. 

\noindent Compared to the meta-transfer learning-based baggage threat detector (TSD) \cite{hassan2020Sensors}, our framework  achieves 15.62\% higher performance in terms of $\mu_{ap}^{b:50}$ on the SIXray  dataset (Table \ref{tab:gd1}). However, on GDXray \cite{mery2015gdxray}, it lags from the TSD by 6.61\%. The superiority of \cite{hassan2020Sensors} here stems from its capacity to generate the dual-energy tensors \cite{hassan2020Sensors} that can effectively highlight the transitions of the contraband items from the grayscale X-ray scans. However, TSD is still sensitive to extremely cluttered baggage threats, as evident through its performance on the SIXray \cite{miao2019sixray} dataset.

}
\noindent The performance of CIE-Net, in terms of the $\mu_{ap}^{b:50}$, is although lagging from the original CST framework \cite{hassan2019} in Table \ref{tab:gd1}. But this comparison is unfair as the original CST \cite{hassan2019} framework is non-incremental and uses more training data and computational resources to produce these results. Nevertheless, under fair comparison, the CIE-Net outperforms CST \cite{hassan2019} by 4.52\% and 3.94\% on GDXray \cite{mery2015gdxray} and SIXray \cite{miao2019sixray}, respectively, in $\mu_{ap}^{b:50}$ (see Table \ref{tab:gd1}). 
Also, the CST framework is extremely parametric dependent (i.e., it has to be tuned for each dataset independently). Therefore, it does not generalize well for scans and datasets having drastically varying properties. Furthermore, it also lacks the inherent ability to generate items mask and falls under conventional object detectors.  

\T{\noindent With regard to run-time performance, the CIE-Net is about two-time faster than several instance segmentation models like MS R-CNN  \cite{msrcnn}, Mask R-CNN \cite{maskrcnn}, and HTC \cite{htc}. It also showed a modest performance compared to  YOLOv3 \cite{yolov3},  CST \cite{hassan2019},  RetinaNet \cite{retinanet}  and  YOLCAT  \cite{yolact}. Nonetheless, looking at both accuracy and efficiency figures in, respectively,  Table \ref{tab:iou}, \ref{tab:gd1}, and \ref{tab:time}, we can assert that the CIE-Net realizes the best trade-off between time and performance. It is also important to point out that the CIE-Net model's current conception is mainly driven by accurately recognizing the \RV{cluttered and overlapping} contraband items rather than achieving efficiency. However, we envisage different measures to enhance this aspect in the future. A first remedy can be replacing the conventional convolutional blocks with residual driven atrous convolutions (with variable dilation factors) \cite{ragnetv2, Wang2018WACV},  resulting in a significant reduction of the trainable parameters, thus increasing the overall run-time performance by many folds. Furthermore, we can generate a lightweight version of the CIE-Net by employing a switching mechanism \cite{chen2019Access} to process only positive regions showcasing contraband items and their instances while ignoring the negative regions. \RV{In addition to this, we also envisage employing multi-task attention networks \cite{wang2020MAN} and adversarial domain adaptation \cite{wang2019ADA} schemes as future work to further improve the threat detection performance of the proposed framework.}

}

\section{Conclusion} \label{sec:conclusion}

\T{\noindent This paper presents a novel instance segmentation framework that utilizes incremental learning and a conventional encoder-decoder architecture to extract and recognize heavily cluttered, occluded, and overlapping contraband items from multi-vendor baggage X-ray scans. 
Since the proposed framework is powered through incremental learning, it reaps the benefit of using small-scale training data and bypasses hectic ground-truth generation mechanisms to make semantic segmentation networks instance-aware. The proposed framework has an in-built capacity to resist catastrophic forgetting through a proposed instance segmentation loss function, introducing a  novel feature of incorporating mutual information loss embedding the complex inter-dependencies between old knowledge and newly learned information through Bayesian inference. The proposed scheme is unique as it modifies the conventional semantic segmentation networks to perform instance-aware segmentation via incremental learning. 
\noindent By being trained on two different datasets and their combination, the proposed framework produces the best results compared to existing state-of-the-art solutions in multiple metrics, evidencing the ability to effectively recognize the cluttered and \RV{overlapping} objects through instance segmentation rather than through object detectors. To the best of our knowledge, it is the only framework to date, which can accurately extract overlapping baggage items from the multi-vendor grayscale and colored X-ray images (in an incremental fashion) despite the significant variations in the scan features of both datasets. 
\noindent In addition to the envisaged task mentioned in the Discussion section to optimize the model design, future work will consider investigating the challenging problem of detecting 3D-printed items (e.g., \textit{guns}).  These items, made from organic matter, have low visibility in the  X-ray scans. Devising proper models for this category of objects is our potential future work. 
}

\small
\bibliographystyle{ieeetr}
\bibliography{main.bib}

\end{document}